\documentclass[11pt,a4paper]{article}
\usepackage{times,latexsym}
\usepackage{url}
\usepackage[T1]{fontenc}

\usepackage[acceptedWithA]{tacl2021v1}
\usepackage{xspace,mfirstuc,tabulary}

\newif\iftaclinstructions
\taclinstructionsfalse % AUTHORS: do NOT set this to true
\iftaclinstructions
\renewcommand{\confidential}{}
\renewcommand{\anonsubtext}{(No author info supplied here, for consistency with
TACL-submission anonymization requirements)}
\newcommand{\instr}
\fi

\iftaclpubformat % this "if" is set by the choice of options

\else

\fi

\usepackage{hyperref}
\usepackage{times}
\usepackage{latexsym}

\usepackage[T1]{fontenc}
\usepackage[utf8]{inputenc}

\usepackage{microtype}

\usepackage{multirow}
\usepackage{booktabs}
\usepackage{amsmath}
\usepackage{graphics}
\usepackage{graphicx}

\usepackage{caption}
\usepackage{subcaption}
\usepackage{tabularx}

\usepackage{makecell}
\usepackage{color, colortbl}
\definecolor{Gray}{gray}{0.9}
\definecolor{DarkGray}{gray}{0.7}

\usepackage{cleveref}
\crefformat{section}{\S#2#1#3}
\crefformat{subsection}{\S#2#1#3}
\crefformat{subsubsection}{\S#2#1#3}
\crefrangeformat{section}{\S\S#3#1#4 to~#5#2#6}
\crefmultiformat{section}{\S\S#2#1#3}{ and~#2#1#3}{, #2#1#3}{ and~#2#1#3}
\usepackage{refstyle}
\Crefformat{figure}{#2Fig.~#1#3}
\Crefmultiformat{figure}{Figs.~#2#1#3}{ and~#2#1#3}{, #2#1#3}{ and~#2#1#3}
\Crefformat{table}{#2Table~#1#3}
\Crefmultiformat{table}{Tabs.~#2#1#3}{ and~#2#1#3}{, #2#1#3}{ and~#2#1#3}
\Crefformat{appendix}{Appx.~\S#2#1#3}
\crefformat{algorithm}{Alg.~#2#1#3}

\title{Visual Spatial Reasoning}

\author{Fangyu Liu \ \ \ \ \ \ \ \  Guy Emerson \ \ \ \ \ \ \ \ Nigel Collier \\
        University of Cambridge \\
        \texttt{\{fl399, gete2, nhc30\}@cam.ac.uk}}

\begin{document}
\maketitle

\begin{abstract}
Spatial relations are
%fundamental
a basic part of human cognition.
%and are the most basic knowledge for us to understand and communicate about our physical surroundings.
%In this paper, we ask the critical question: Are current vision-and-language models (VLMs) able to correctly understand spatial relations? And how well their judgements align with human?
%To answer this question, we propose Visual Spatial Reasoning (VSR), a novel human labelled dataset for investigating VLMs' capabilities and also study the linguistic challenges in spatial recognition.
However, they are expressed in natural language in a variety of ways,
and previous work has suggested that current vision-and-language models (VLMs) struggle to capture relational information.
In this paper, we present Visual Spatial Reasoning (VSR),
a dataset containing more than 10k natural text-image pairs with 66 types of spatial relations in English (such as: under, in front of, facing).
%Specifically, given a caption and an image, the model needs to perform binary classification and decide if the caption accurately describes the spatial relation of two objects presented in the image. 
%While being seemingly simple and straightforward, the dataset has complex linguistic nature since the use of language can be versatile when expressing a visual scene. 
%The linguistic diversity is reflected in aspects such as the choice of frames of reference.
While using a seemingly simple annotation format, we show how the dataset includes challenging linguistic phenomena, such as varying reference frames.
%With comprehensive experiments,
We demonstrate a large gap between human and model performance: the human ceiling is above 95\%, while state-of-the-art models only achieve around 70\%.
%With fine-grained categorisation and control on both concepts and relations, our VSR benchmark enables us to perform interesting probing analysis to pinpoint VLMs' failure cases and the reasons behind. 
We observe that VLMs' by-relation performances have little correlation with the number of training examples and the tested models are in general incapable of recognising relations concerning the orientations of objects.%
%\footnote{We will publicly release the dataset and code.}
%Also, VLMs have poor zero-shot generalisation toward unseen concepts. 
%Besides highlighting the deficiencies of VLMs, VSR is also a valuable linguistic resource for studying human variance in perceiving spatial relations where we pay special attention to prominent features such as frame of reference used by the annotators.
%
\footnote{Data and code: \href{https://github.com/cambridgeltl/visual-spatial-reasoning}{github.com/cambridgeltl/visual-spatial-reasoning}.}
\end{abstract}

\section{Introduction}\label{sec:intro}

Multimodal NLP research has developed rapidly in recent years, with substantial performance gains on tasks such as
%Multi-modal NLP research has been developing rapidly in the last couple of years with the introduction of %a series of
%pre-trained vision-language models (VLMs) such as ViLBERT \citep{lu2019vilbert}, VisualBERT \citep{li2019visualbert}, LXMERT \citep{tan2019lxmert}, UNITER \citep{chen2020uniter}, and ViLT \citep{kim2021vilt}, 
%to name a few. VLMs have pushed state-of-the-art performance up by
%significant
%substantial margins on vision-language benchmarks, including tasks such as 
visual question answering (VQA) \citep{antol2015vqa,johnson2017clevr,goyal2017making,hudson2019gqa,zellers2019recognition}, vision-language reasoning or entailment \citep{suhr-etal-2017-corpus,suhr-etal-2019-corpus,xie2019visual,liu-etal-2021-visually}, and referring expression comprehension \citep{yu2016modeling,liu2019clevr}. 
Existing benchmarks, such as NLVR2 \citep{suhr-etal-2019-corpus} and VQA \citep{goyal2017making}, define generic paradigms for testing vision-language models (VLMs). However, as we further discuss in \Cref{sec:rw}, these benchmarks are not ideal for probing VLMs as they typically conflate multiple sources of error and do not allow controlled analysis on specific linguistic or cognitive properties, making it difficult to categorise and fully understand the model failures.
In particular, spatial reasoning has been found to be particularly challenging for current models, and much more challenging than capturing properties of individual entities \citep{kuhnle-etal-2018-clever,cirik-etal-2018-visual,akula-etal-2020-words}, even for state-of-the-art models such as CLIP \citep{radford2021learning,subramanian2022reclip}.

% \begin{figure}[!tbp]
%   \centering
%     \includegraphics[width=0.8\linewidth]{images/000000259555.jpg}
%     \caption{Caption: \textit{The potted plant is at the right side of the bench}. Label: \texttt{True}. 
%     %All models predicted \texttt{False}.
%     }
%     \label{fig:potted_plant_bench}
%     \vspace{-1em}
% \end{figure}

Another line of work generates synthetic datasets in a controlled manner to target specific relations and properties when testing VLMs, e.g., CLEVR \citep{liu2019clevr} and ShapeWorld \citep{kuhnle-copestake-2018-deep}. 
However, synthetic datasets may accidentally overlook challenges (such as orientations of objects which we will discuss in \Cref{sec:exp}), and using natural images allows us to explore a wider range of language use.
%But we will demonstrate later in our paper that synthetic datasets neglect many realistic challenges presented in natural images. 

To address the lack of probing evaluation benchmarks in this field, we present VSR (Visual Spatial Reasoning), a controlled dataset that explicitly tests VLMs for spatial reasoning. We choose spatial reasoning as the focus because it is one of the most fundamental capabilities for both humans and VLMs. 
Such relations are crucial to how humans organise their mental space and make sense of the physical world, and therefore fundamental for a grounded semantic model \citep{talmy1983space}.

%When humans process a visual scene, different \emph{frames of reference} can be used, further adding complexity to the task. Specifically, frame of reference is a coordinate system used to identify the physical location of an object.
%In data collection, we found that reference frame use by humans can vary when objects presented in the image differ.

\begin{figure}[!t]
  \centering
     \includegraphics[width=0.86\linewidth]{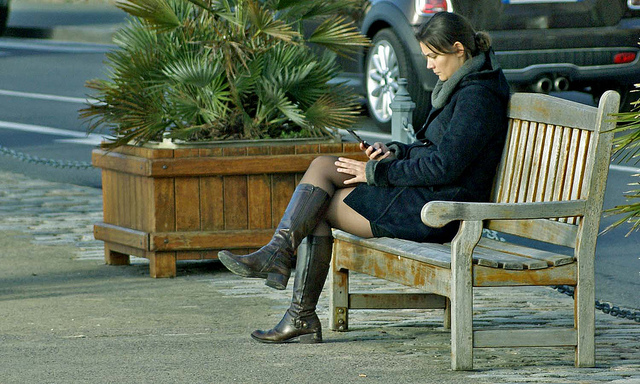}
    \caption{Caption: \textit{The potted plant is at the right side of the bench}. Label: \texttt{True}.}
    \label{fig:potted_plant_bench}
    % \includegraphics[width=0.9\linewidth]{images/000000292365.jpg}
    % \caption{Caption: \textit{The cat is inside the toilet.} Label: \texttt{False}.}
    % \label{fig:cat_toilet}
    \vspace{\floatsep}
        \includegraphics[width=0.86\linewidth]{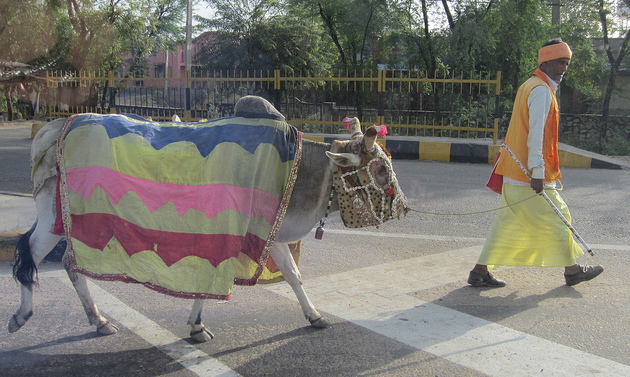}
    \caption{Caption: \textit{The cow is ahead of the person.} Label: \texttt{False}.}
     \label{fig:person_cow}
\end{figure}

The VSR dataset contains natural image-text pairs in English,
with the data collection process explained in~\Cref{sec:dataset_create}.
Each example in the dataset consists of an image and a natural language description which states a spatial relation of two objects presented in the image (two examples are shown in \Cref{fig:potted_plant_bench} and \Cref{fig:person_cow}). 
A VLM needs to classify the image-caption pair as either true or false, indicating whether the caption is correctly describing the spatial relation. 
The dataset covers 66 spatial relations and has >10k data points, using 6,940 images from MS COCO \citep{lin2014microsoft}. 

Situating one object in relation to another requires a \textit{frame of reference}: a system of coordinates against which the objects can be placed.
%\footnote{We have two design choices...}
Drawing on detailed studies of more than forty typologically diverse languages, \citet{levinson2003space} concludes that the diversity can be reduced to three major types: intrinsic, relative, and absolute.
An intrinsic frame is centred on an object, e.g., \textit{behind the chair}, meaning at the side with the backrest.
A relative frame is centred on a viewer, e.g., \textit{behind the chair}, meaning further away from someone's perspective.
An absolute frame uses fixed coordinates, e.g., \textit{north of the chair},
using cardinal directions.
In English, absolute frames are rarely used when describing relations on a small scale, and they do not appear in our dataset.
However, intrinsic and relative frames are widely used, and present an important source of variation.
We discuss the impact on data collection in~\Cref{sec:human_val},
and analyse the collected data in~\Cref{sec:dataset_analysis}.

We test four popular VLMs, i.e., VisualBERT \citep{li2019visualbert}, LXMERT \citep{tan2019lxmert}, ViLT \citep{kim2021vilt}, and CLIP \citep{radford2021learning} on VSR,
with results given in~\Cref{sec:exp}. While the human ceiling is above 95\%, all four models struggle to reach 70\% accuracy. We conduct comprehensive analysis on the failures of the investigated VLMs and highlight that (1) positional encodings are extremely important for the VSR task; (2) models' by-relation performance barely correlates with the number of training examples; (3) in fact, several spatial relations that concern orientation of objects are especially challenging for current VLMs; (4) VLMs have extremely poor generalisation on unseen concepts.
%\footnote{Objects that do not appear in training set but appears in test set.}.

% Why not currently existing dataset? (e.g. NLVR2, CLEVR) 

\section{Related Work}\label{sec:rw}

%\begin{itemize}
    % \item Spatial Relation Resolver \url{https://openreview.net/pdf?id=eT83GooycXa}
    % %\item NLVR/NLVR2
    % \item see NLVR2 paper
    % \item Analysis? VL-InterpreT: An Interactive Visualization Tool for Interpreting Vision-Language Transformers \url{https://arxiv.org/abs/2203.17247}
%    \item CLEVR: A Diagnostic Dataset for Compositional Language and Elementary Visual Reasoning
%\end{itemize}

\subsection{Comparison with synthetic datasets}
Synthetic language-vision reasoning datasets, e.g., SHAPES \citep{andreas2016neural}, CLEVR \citep{liu2019clevr}, NLVR \citep{suhr-etal-2017-corpus}, and ShapeWorld \citep{kuhnle-copestake-2018-deep}, enable full control of dataset generation and could potentially benefit probing of spatial reasoning capability of VLMs. They share a similar goal to us, to diagnose and pinpoint weaknesses in VLMs. However, synthetic datasets necessarily simplify the problem as they have inherently bounded expressivity. In CLEVR, objects can only be spatially related via four relationships: ``left'', ``right'', ``behind'', and ``in front of'' while VSR covers 66 relations.

Synthetic data does not always accurately reflect the challenges of reasoning in the real world. For example, objects like spheres, which often appear in synthetic datasets, do not have orientations. In real images, orientations matter and human language use depends on that. Furthermore, synthetic images do not take the scene as a context into account. The interpretation of object relations can depend on such scenes (e.g., the degree of \textit{closeness} can vary in open space and indoor scenes).
% Why not automatically generate a dataset with rules? 

Last but not least, the vast majority of spatial relationships cannot be determined by rules. Even for the seemingly simple relationships like ``left/right of'', the determination of two objects' spatial relationships can depend on the observer's viewpoint, whether the object has a \textit{front}, if so, what are their orientations, etc.

\subsection{Spatial relations in existing vision-language datasets}
Several existing vision-language datasets with natural images also contain spatial relations (e.g., NLVR2, COCO, and VQA datasets). 
\citet{suhr-etal-2019-corpus} summarise that there are 9 prevalent linguistic phenomena/challenges in NLVR2 \citep{suhr-etal-2019-corpus} such as coreference, existential quantifiers, hard cardinality, spatial relations, etc., and 4 in VQA datasets \citep{antol2015vqa,hudson2019gqa}. 
However, the different challenges are entangled in these datasets. Sentences contain complex lexical and syntactic information and can thus conflate different sources of error, making it hard to identify the exact challenge and preventing categorised analysis. 
\citet{yatskar-etal-2016-stating} extract 6 types of visual spatial relations directly from MS COCO images with annotated bounding boxes. But rule-based automatic extraction can be restrictive as most relations are complex and cannot be identified relying on bounding boxes.
Recently, \citet{positional2022} extract captions that contain 28 positional keywords from MS COCO and swap the keywords with their antonyms to construct a challenging probing dataset. However, the COCO captions also have the error-conflation problem. Also, the number of examples and types of relations are restricted by COCO captions.
%\item Probing the Role of Positional Information in Vision-Language Models (\url{https://openreview.net/pdf?id=2Ubik08ztdB})

Visual Genome \citep{krishna2017visual} also contains annotations of objects' relations including spatial relations. However, it is only a collection of true statements and contains no negative ones, so cannot be framed as a binary classification task.
It is non-trivial to automatically construct negative examples since multiple relations can be plausible for a pair of object in a given image.
Relation classifiers are harder to learn than object classifiers on this dataset \citep{liu-emerson-2022-learning}.

\citet{parcalabescu2021valse} propose a benchmark called VALSE for testing VLMs' capabilities on various linguistic phenomena. VALSE has a subset focusing on ``relations'' between objects. It uses texts modified from COCO's original captions. However, it is a zero-shot benchmark without training set, containing just 535 data points. So, it is not ideal for large-scale probing on a wide spectrum of spatial relations.

\subsection{Spatial reasoning without grounding}
There has also been interest in probing models' spatial reasoning capability without visual input. For example,
\citet{collell2018acquiring,mirzaee-etal-2021-spartqa,liu2022things} probe pretrained text-only models or VLMs' spatial reasoning capabilities with text-only questions.
However, a text-only dataset cannot evaluate how a model relates language to grounded spatial information.
In contrast, VSR focuses on the joint understanding of vision and language input.

\subsection{Spatial reasoning as a sub-component}
%TODO: add additional RW 
Last but not least, some vision-language tasks and models require spatial reasoning as a sub-component.
For example, \citet{lei-etal-2020-tvqa} propose TVQA+, a spatio-temporal video QA dataset containing bounding boxes for objects referred in the questions. Models then need to simultaneously conduct QA while detecting the correct object of interest. \citet{christie-etal-2016-resolving} propose a method for simultaneous image segmentation and prepositional phrase attachment resolution. Models have to reason about objects' spatial relations in the visual scene to determine the assignment of prepositional phrases.
However, if spatial reasoning is only a sub-component of a task, error analysis becomes more difficult.
In contrast, VSR provides a focused evaluation of spatial relations, which are particularly challenging for current models.

\section{Dataset Creation}\label{sec:dataset_create}

In this section we detail how VSR is constructed. The data collection process can generally be split into two phases (1) contrastive caption generation (\Cref{sec:cap_gen}) and (2) second-round validation (\Cref{sec:human_val}). We then discuss annotator hiring \& payment (\Cref{sec:annotator}), dataset splits (\Cref{sec:splits}), and the human ceiling \& agreement of VSR (\Cref{sec:human_ceil}).

\begin{figure}[!tbp]
  \centering
    \includegraphics[width=0.99\linewidth]{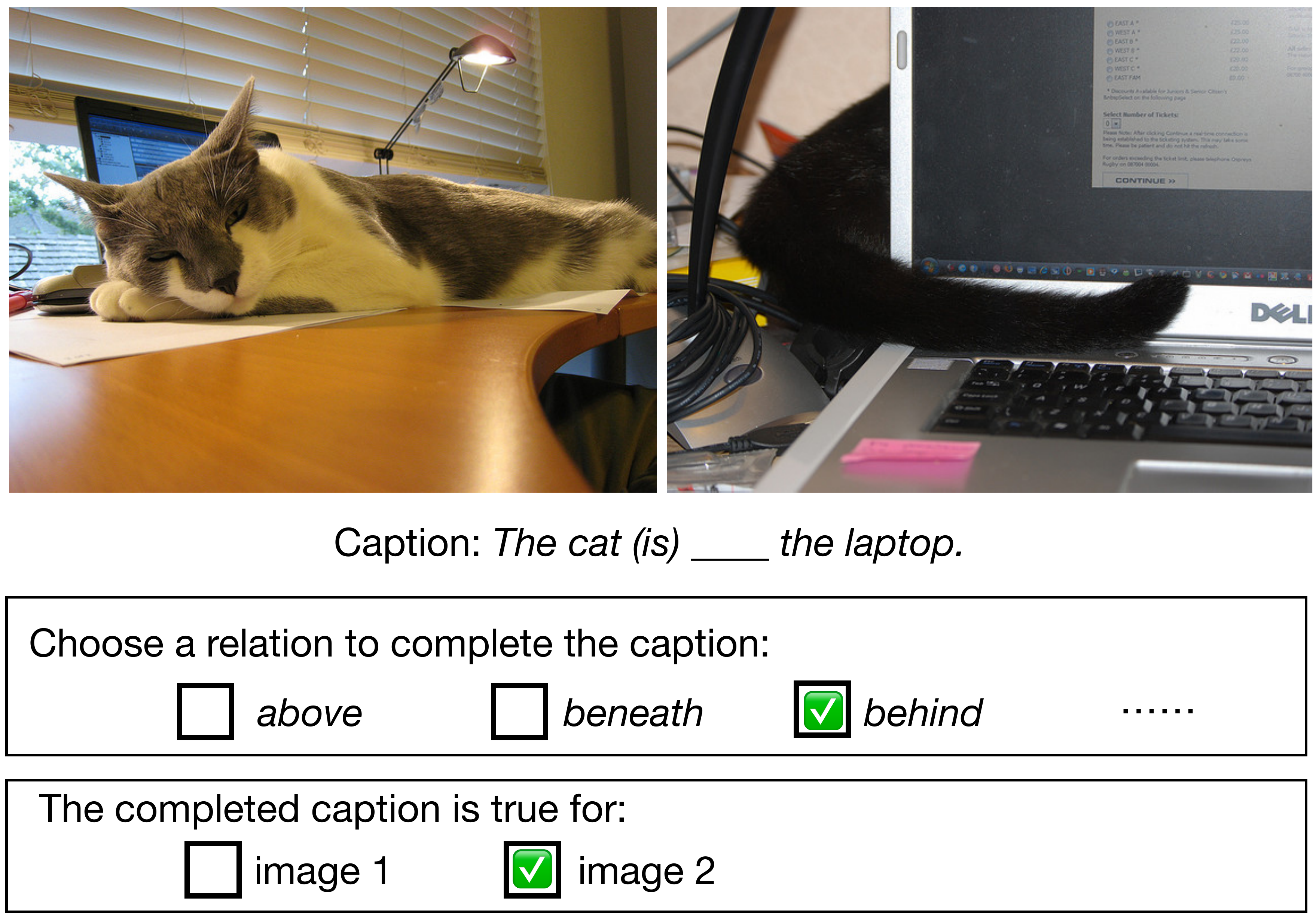}
    \caption{An annotation example of concepts ``cat'' \& ``laptop'' in contrastive caption generation. The example generates two data points for our dataset: one ``True'' instance when the completed caption is paired with image 2 (right) and one ``False'' instance when paired with image 1 (left).}
    \label{fig:annotation_instance}
    %\vspace{-1em}
\end{figure}

\subsection{Contrastive Template-based Caption Generation (\Cref{fig:annotation_instance})}\label{sec:cap_gen}

In order to highlight spatial relations and avoid annotators frequently choosing trivial relations (such as ``near to''), we use a contrastive caption generation approach. Specifically, first, a pair of images, each containing two concepts of interests, would be randomly sampled from MS COCO (we use the train and validation sets of COCO 2017). Second, an annotator would be given a template containing the two concepts and is required to choose a spatial relation from a pre-defined list (\Cref{tab:spatial_relations}) that makes the caption correct for one image but incorrect for the other image. We will detail these steps and explain the rationales in the following. 

\paragraph{Image pair sampling.} MS COCO 2017 contains 123,287 images and has labelled the segmentation and classes of 886,284 instances (individual objects). Leveraging the segmentation, we first randomly select two concepts (e.g., ``cat'' and ``laptop'' in \Cref{fig:annotation_instance}), then retrieve all images containing the two concepts in COCO 2017 (train and validation sets). Then images that contain multiple instances of any of the concept are filtered out to avoid referencing ambiguity. For the single-instance images, we also filter out any of the images with instance pixel area size $<30,000$, to prevent extremely small instances. After these filtering steps, we randomly sample a pair in the remaining images. We repeat such a process to obtain a large number of individual image pairs for caption generation.

\begin{table*}[tbp]
\small
    \centering
    \begin{tabular}{l|l}
    \toprule
    \rowcolor{DarkGray}
    Category & Spatial Relations \\
    \midrule
    Adjacency   & \makecell[l]{Adjacent to, alongside, at the side of, at the right side of, at the left side of, attached to,\\ at the back of, ahead of, against, at the edge of} \\
    \rowcolor{Gray}
 Directional & \makecell[l]{Off, past, toward, down, deep down$^\ast$, up$^\ast$, away from, along, around, from$^\ast$, into, to$^\ast$,\\ across, across from, through, down from }\\
    Orientation & Facing, facing away from, parallel to, perpendicular to\\
    \rowcolor{Gray}
    Projective & \makecell[l]{On top of, beneath, beside, behind, left of, right of, under, in front of, below, above, over,\\ in the middle of}\\
    Proximity & By, close to, near, far from, far away from \\
    \rowcolor{Gray}
    Topological & \makecell[l]{Connected to, detached from, has as a part, part of, contains, within, at, on, in, with,\\ surrounding, among, consists of, out of, between, inside, outside, touching}\\
    Unallocated & Beyond, next to, opposite to, after$^\ast$, among, enclosed by \\
\bottomrule
    \end{tabular}
    \caption{The 71 available spatial relations. 66 of them appear in our final dataset ($\ast$ indicates not used).}
    % unsued: 'up', 'through', 'deep down', 'from', 'to', 'after'
    \label{tab:spatial_relations}
\end{table*}

\paragraph{Fill in the blank: template-based caption generation.}
Given a pair of images, the annotator needs to come up with a valid caption that makes it a correct description for one image but incorrect for the other. In this way, the annotator should focus on the key difference between the two images (which should be a spatial relation between the two objects of interest) and choose a caption that differentiates the two. Similar paradigms are also used in the annotation of previous vision-language reasoning datasets such as NLVR(2) \citep{suhr-etal-2017-corpus,suhr-etal-2019-corpus} and MaRVL \citep{liu-etal-2021-visually}. To regularise annotators from writing modifiers and differentiating the image pair with things beyond accurate spatial relations, we opt for a template-based classification task instead of free-form caption writing.\footnote{\citet{hendricks-nematzadeh-2021-probing} propose a zero-shot probing benchmark of similar spirit for \textit{verb} understanding. All captions are simplified as subject-verb-object triplets.}
 Besides, the template-generated dataset can be easily categorised based on relations and their categories. Specifically, the annotator would be given instance pairs as shown in \Cref{fig:annotation_instance}.
 %\footnote{The actual interface for annotation is in Appendix \Cref{fig:annotation_interface}.}
 
The caption template has the format of  ``\textit{The \texttt{ENT1} (is) \rule{1cm}{0.15mm} the \texttt{ENT2}.}'', and the annotators are instructed to select a relation from a fixed set to fill in the slot. The copula ``is'' can be omitted for grammaticality. For example, for ``contains'' %``consists of'', 
and ``has as a part'', ``is'' should be discarded in the template when extracting the final caption. 

% we use 
% ``\textit{The \texttt{OBJ1} \rule{1cm}{0.15mm} the \texttt{OBJ2}.}'' to generate the final captions \Cref{fig:annotation_instance} has a bracket around ``is'' since ``\textit{The \texttt{OBJ1} (is) \rule{1cm}{0.15mm} the \texttt{OBJ2}.}'' is not compatible with all relations. For ``contains'', ``consists of'', and ``has as a part'', we use 
% ``\textit{The \texttt{OBJ1} \rule{1cm}{0.15mm} the \texttt{OBJ2}.}'' to generate the final captions, ensuring the sentences being grammatical.

The fixed set of spatial relations enable us to obtain the full control of the generation process. The full list of used relations are listed in \Cref{tab:spatial_relations}. It contains 71 spatial relations and is adapted from the summarised relation table of \citet{marchi2021cross}. We made minor changes to filter out clearly unusable relations, made relation names grammatical under our template, and reduced repeated relations.
%\footnote{The details of each adaptation is listed in Appendix, \Cref{tab:spatial_relations_adaptations}.}
In our final dataset, 66 out of the 71 available relations are actually included (the other 6 are either not selected by annotators or are selected but the captions did not pass the validation phase).

\subsection{Second-round Human Validation}\label{sec:human_val}

% \begin{figure}[!tbp]
%   \centering
%     \includegraphics[width=0.6\linewidth]{images/000000434261.jpg}
%     \caption{Caption: \textit{The person is at the left side of the bus}. Label: \texttt{True}.}
%     \label{fig:frame_example}
% \end{figure}

\begin{figure}[!tbp]
  \centering
    \includegraphics[width=0.88\linewidth]{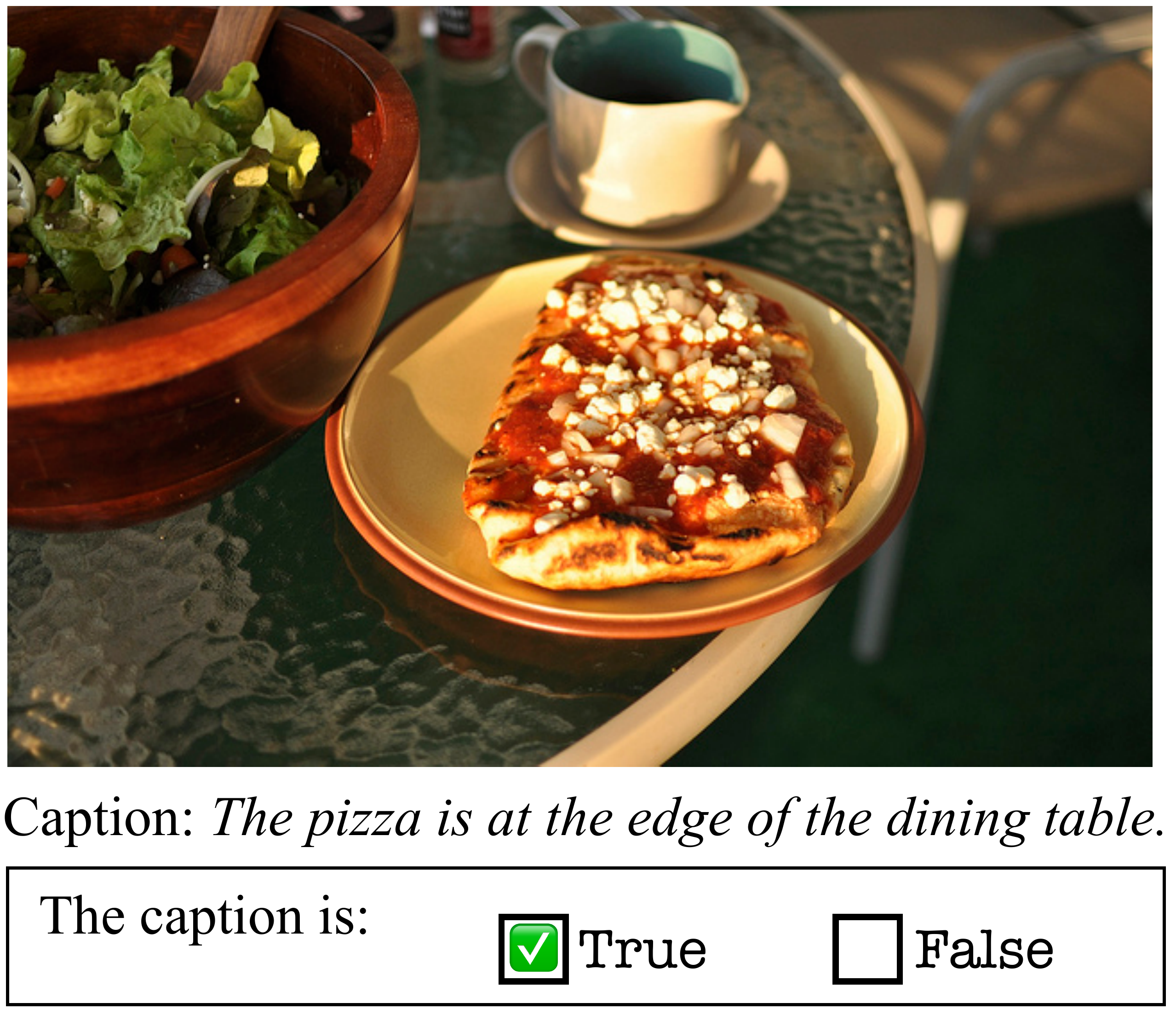}
    \caption{A second-round validation example.}
    %[TODO]: fix typo, i.e. dinning}
    \label{fig:val_example}
    %\vspace{-1em}
\end{figure}

In the second-round validation, every annotated data point is reviewed by at least 3 additional human annotators (validators).
Given a data point (consisting of an image and a caption), the validator gives either a \texttt{True} or \texttt{False} label as shown in \Cref{fig:val_example} (the original label is hidden).
%\footnote{The actual validation interface is in Appendix \Cref{fig:validation_interface}.} 
%As mentioned above, we instruct validators to give \texttt{True} label to statements that are feasible under at least one reference frame. 
%We exclude data points that have $<\frac{2}{3}$ validators agreeing with the original label.
In our final dataset, we exclude instances with fewer than 2 validators agreeing with the original label.

\paragraph{Design choice on reference frames.} During validation, a validator needs to decide whether a statement is true or false for an image. However, as discussed in~\Cref{sec:intro}, interpreting a spatial relation requires choosing a \textit{frame of reference}.
For some images, a statement can be both true and false, depending on the choice.
%(or \textit{reference frame})  \citep{levinson2003space}.
%\citet{levinson2003space} summarise that there are three possible reference frames: (1) intrinsic (coordinates centred on an object: ``\textit{the cat is in front of the house}''); (2) relative (coordinates centred on a viewpoint: ``\textit{the cat is to the left of the house}''), and (3) absolute (fixed coordinates: ``\textit{the cat is in the north of the house}''). In our work, we found that intrinsic and relative are the most used in English while absolute is very rarely used so we do not consider it.
As a concrete example, in \Cref{fig:potted_plant_bench}, while the potted plant is on the left side from the viewer's perspective (relative frame), the potted plant is at the right side if the bench is used to define the coordinate system (intrinsic frame).

In order to ensure that annotations are consistent across the dataset,
we communicated to the annotators that, for relations such as ``left''/``right'' and ``in front of''/``behind'', they should consider both possible reference frames, and assign the label \texttt{True} when a caption is true from either the intrinsic or the relative frame.
% \footnote{In other words, under our labelling scheme, both \texttt{True} and \texttt{False} labels for this specific example is correct.} 
Only when a caption is incorrect under both reference frames (e.g., if the caption is ``\textit{The potted plant is under the bench.}'' for \Cref{fig:potted_plant_bench}) should a \texttt{False} label be assigned.

On a practical level, this adds difficulty to the task, since a model cannot naively rely on pixel locations of the objects in the images, but also needs to correctly identify orientations of objects.
However, the task is well-defined:
a model that can correctly simulate both reference frames would be able to perfectly solve this task.

From a theoretical perspective, by involving more diverse reference frames, we are also demonstrating the complexity of human cognitive processes when understanding a scene, since different people approach a scene with different frames.
Attempting to enforce a specific reference frame would be methodologically difficult and result in an unnaturally restricted dataset.

\begin{figure*}[tbh]
    \centering
    \includegraphics[width=\linewidth]{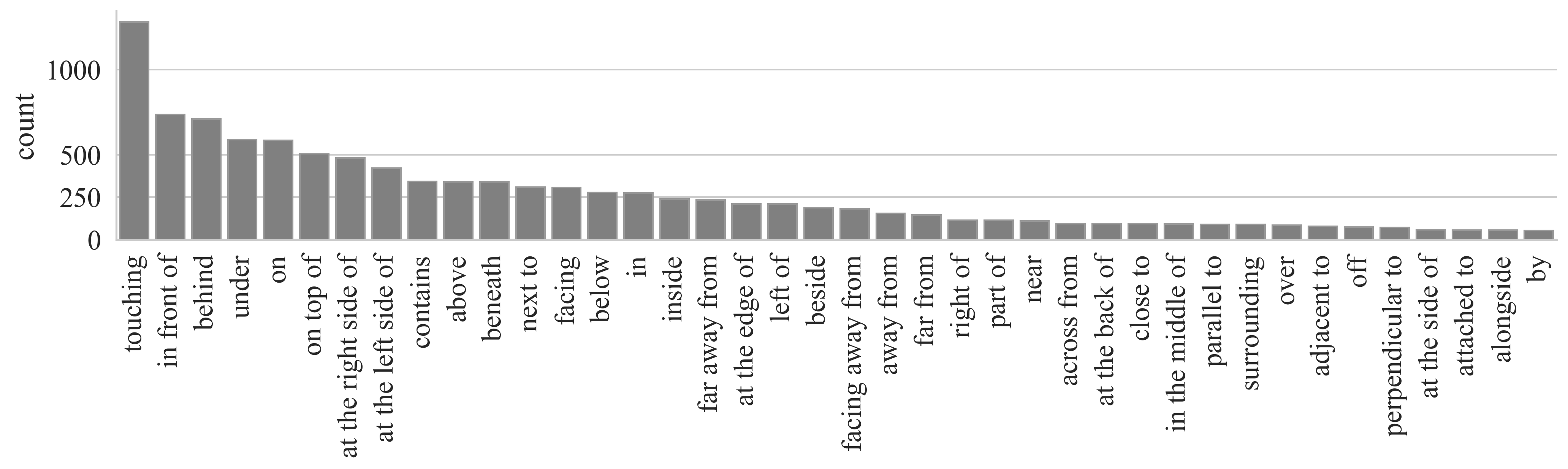}
\caption{Relation distribution of the final dataset (sorted by frequency). Top 40 most frequent relations are included. 
%See Appendix \Cref{table:rel_stats_full} for the full listing. 
It is clear that the relations follow a long-tailed distribution.}
%(however, according to our sample efficiency analysis, it is not the frequency of training examples causing the major challenge).}
    \label{fig:rel_dist}
\end{figure*}

\begin{figure*}[tbh]
    \centering
    \includegraphics[width=\linewidth]{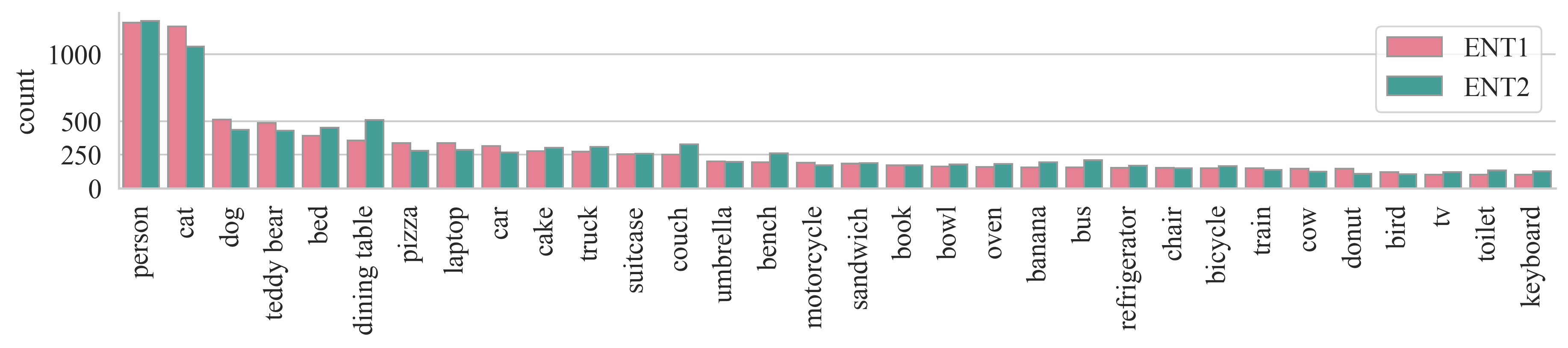}
%\caption{Concept distribution. Only concepts with $>100$ frequencies at subject and object positions are included in the figure.}
\caption{Concept distribution. Only concepts with $>100$ frequencies are included.}
%\vspace{-1em}
    \label{fig:concept_dist}
\end{figure*}

% \begin{table}
% \small
% \centering
% \begin{tabular}{llll}
% \toprule
%  avg. \# validation & \# 100\% agree & \# $\ge2/3$ consensus \\
%  \midrule
%  2.29 & 9,384 (73.2\%) & 9,958 (77.7\%)\\
% \bottomrule
% \end{tabular}
% \caption{Validation phase statistics. Number of total raw data points is 12,809.}
% \label{table:val_stats}
% \end{table}

\subsection{Annotator Hiring and Organisation}\label{sec:annotator}

Annotators were hired from \url{prolific.co}. We required them to (1) have at least a bachelor's degree, (2) be fluent in English, and (3) have a $>$99\% historical approval rate on the platform. All annotators were paid 12 GBP per hour. %Prolific takes an extra 33\% of service charge and 20\% VAT on the service charge. % not important and to add back later

For caption generation, we released the task with batches of 200 instances and the annotator was required to finish a batch in 80 minutes. An annotator could not take more than one batch per day. In this way we had a diverse set of annotators and could also prevent annotators from becoming fatigued. For second-round validation, we grouped 500 data points in one batch and an annotator was asked to label each batch in 90 minutes.

In total, 24 annotators participated in caption generation and 45 participated in validation. 4 people participated in both phases, which should have minimally impacted the validation quality. The annotators had diverse demographic backgrounds: they were born in 15 countries, were living in 13 countries, and had 12 nationalities. 50 annotators were born and living in the same country while others had moved to different ones. The vast majority of our annotators were residing in the UK (32), South Africa (9), and Ireland (7).
The ratio for holding a Bachelor/Master/PhD as the highest degree was: 12.5\%/76.6\%/10.9\%. 
Only 7 annotators were non-native English speakers while the other 58 were native speakers.
56.7\% of the annotators self-identified as female and 43.3\% as male.

\subsection{Dataset Splits}\label{sec:splits}
We split the 10,972 validated data points into train/dev/test sets in two different ways.
The stats of the two splits are shown in \Cref{table:data_final}. In the following, we explain how they are created. 
\textbf{\textit{Random split}:} We split the dataset randomly into train/dev/test with a ratio of 70/10/20. \textbf{\textit{Concept zero-shot split}:}
We create another concept zero-shot split where train/dev/test have no overlapping concepts. I.e., if ``dog'' appears in the train set, then it does not appear in dev or test sets.
This is done by randomly grouping concepts into three sets with a ratio of 50/20/30 of all concepts. This reduces the dataset size, since data poins involving concepts from different parts of the train/dev/test split must be filtered out.
The concept zero-shot split is a more challenging setup since the model has to learn concepts and the relations in a compositional way instead of remembering the co-occurrence statistics of the two.

\begin{table}
\small
\centering
\begin{tabular}{lllll}
\toprule
 split & train & dev & test & total   \\
\midrule
\textit{random} & 7,680 & 1,097 & 2,195 & 10,972 \\
\textit{zero-shot} & 4,713 & 231 & 616 & 5,560\\
\bottomrule
\end{tabular}
\caption{Statistics of the \textit{random} \& \textit{zero-shot} splits.}
\label{table:data_final}
%\vspace{-1em}
\end{table}

%\subsection{Human Ceiling and Inter-annotator Agreement}\label{sec:human_ceil}
\subsection{Human Ceiling and Agreement}\label{sec:human_ceil}

We randomly sample 500 data points from the final random split test set of the dataset for computing human ceiling and inter-annotator agreement.
We hide the labels of the 500 examples and two additional annotators are asked to label \texttt{True}/\texttt{False} for them. On average, the two annotators achieve an accuracy of 95.4\% on the VSR task. We further compute the Fleiss' kappa among the original annotation and the predictions of the two human. The Fleiss' kappa score is 0.895, indicating near-perfect agreement according to \citet{landis1977measurement}.

\section{Dataset Analysis}\label{sec:dataset_analysis}

In this section we compute some basic statistics of our collected data (\Cref{sec:dataset_stats}),
analyse where human annotators have agreed/disagreed (\Cref{sec:agree_disagree}),
and present a case study on reference frames (\Cref{sec:relative_vs_intrinsic}).

\begin{figure*}[t]
    \centering
    \includegraphics[width=\linewidth]{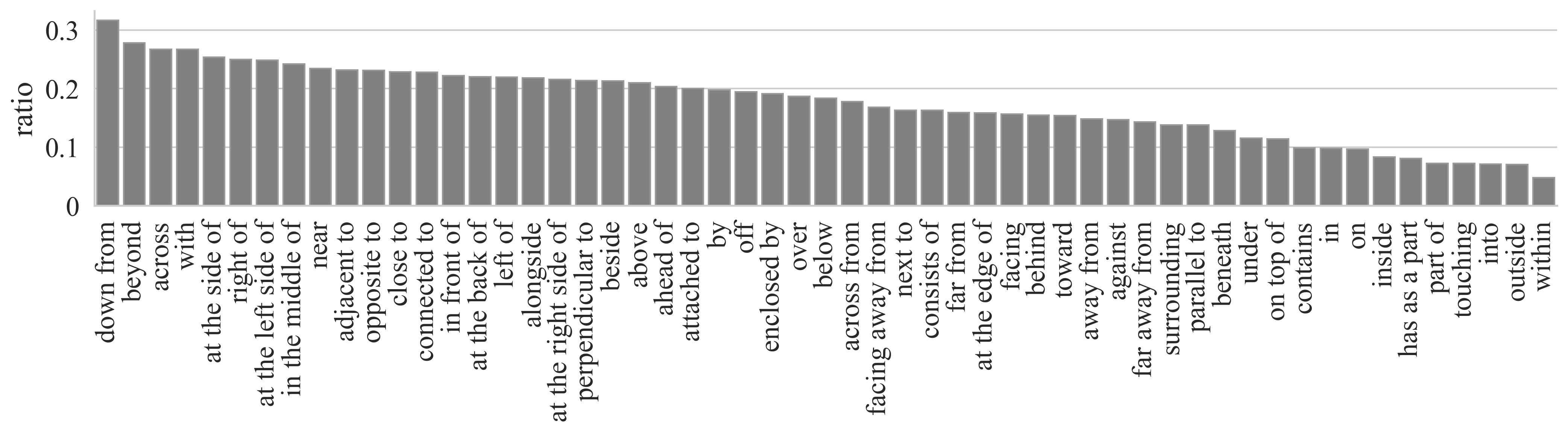}
\caption{Per-relation probability of having two randomly chosen annotator disagreeing with each other (sorted from high to low). Only relations with $>20$ data points are included in the figure.}
%\vspace{-1em}
    \label{fig:disagree_dist}
\end{figure*}

\subsection{Basic Statistics of VSR}\label{sec:dataset_stats}

After the first phase of contrastive template-based caption generation (\Cref{sec:cap_gen}), we collected 12,809 raw data points. 
In the phase of the second round validation (\Cref{sec:human_val}), we collected 39,507 validation labels. 
%On average, every data point received 2.41 validation labels and each data point has at least 2 validation labels. 
Every data point received at least 3 validation labels.
In 69.1\% of the data points, all validators agree with the original label. 85.6\% of the data points have at least $\frac{2}{3}$ annotators agreeing with the original label.
%at least $\frac{2}{3}$ of the validations agreeing with the original label. 
We use $\frac{2}{3}$ as the threshold and exclude all instances with lower validation agreement.
%(in other words, 77.7\% of the raw data points from first round annotation are preserved). 
%We exclude instances with fewer than 2 annotators agreeing with the original label in our final dataset.
After excluding other instances,
%After validation,
10, 972 data points remained and are used as our final dataset.
%\footnote{We introduce the data splits for evaluation in \Cref{sec:splits}.}

Here we provide basic statistics of the two components in the VSR captions: the concepts and the relations. \Cref{fig:rel_dist} demonstrates the relation distribution. ``touching'' is most frequently used by annotators.
% since it can easily differentiate two images if objects in one image is ``touching'' each other while they are not in the other image. 
The relations that reflect the most basic relative coordinates of objects are also very frequent, e.g., ``behind'', ``in front of'', ``on'', ``under'', ``at the left/right side of''. \Cref{fig:concept_dist} shows the distribution of concepts in the dataset. Note that the set of concepts is bounded by MS COCO and the distribution also largely follows MS COCO. Animals such as ``cat'', ``dog'', ``person'' are the most frequent. Indoor objects such as ``dining table'' and ``bed'' are also very dominant. In \Cref{fig:concept_dist}, we separate the concepts that appear at \texttt{ENT1} and \texttt{ENT2} positions of the sentence and their distributions are generally similar.

\subsection{Where do annotators disagree?}\label{sec:agree_disagree}

While we propose using data points with high validation agreement for model evaluation and development, the unfiltered dataset is a valuable resource for understanding cognitive and linguistic phenomena.
We sampled 100 examples where annotators disagree,
and found that around 30 of them are caused by annotation errors but the rest are genuinely ambiguous and can be interpreted in different ways. 
%See Appendix for a couple of examples.
This shows a level of intrinsic ambiguity of the task and variation among people.

Along with the validated VSR dataset, we also release the full unfiltered dataset, with annotators' and validators' metadata, as a second version to facilitate linguistic studies. For example, researchers could investigate questions such as where disagreement is more likely to happen and how people from different regions or cultural backgrounds might perceive spatial relations differently.

To illustrate this, the probability of two randomly chosen annotators disagreeing with each other is given for each relation in \Cref{fig:disagree_dist}.
Some of the relations with high disagreement
%are dependent on the reference frame, e.g., ``at the left side of'' and ``right of''.
can be interpreted in the intrinsic reference frame, which requires identifying the orientations of objects,
e.g.~``at the side of'' and ``in front of''.
Other relations have a high level of vagueness,
e.g., for the notion of \emph{closeness}: ``near'' and ``close to''. By contrast, part-whole relations, such as ``has as a part'', ``part of'', and in/out relations such as ``within'', ``into'', ``outside'' and ``inside'' have the least disagreement.

\subsection{Case Study: Reference Frames}\label{sec:relative_vs_intrinsic}

It is known that the relative reference frame is often preferred in English, at least in standard varieties.
%It was recognised before that relative frame is more commonly used in English
For example, \citet{edmonds2012false} compares Standard Australian English and Aboriginal English, as spoken by school children at a school on Croker Island,
investigating the use of the relations ``in front of'' and ``behind'' when describing simple line drawings of a person and a tree.
Speakers of Standard Australian English were found to prefer the relative frame, while speakers of Aboriginal English were found to prefer the intrinsic frame.

Our methodology allows us to investigate reference frame usage across a wide variety of spatial relations, using a wide selection of natural images.
To understand the frequency of annotators using relative vs. intrinsic frames, we label instances' reference frames and study their distributions.
%of reference when exposed to natural images,
The majority of examples that can be interpreted differently under different reference frames are left/right-related relations (i.e.~``left/right of'' and ``at the left/right side of''). We find all left/right-related \emph{true}\footnote{According to our guideline, false statements are interpreted as false under both frames.} statements and classify them into three categories: (1) intrinsic (2) relative and (3) both (the caption is correct under either intrinsic and relative frames of reference). 
%Note that some captions can be interpreted in both reference frames. 
Among the 616 instances, 68 (11\%) and 518 (84\%) use intrinsic and relative frames respectively, 30 (5\%) can be interpreted with both frames. 
Since the vast majority of our annotators were native English speakers (91\%), and all were university-educated, our finding is consistent with previous work suggesting that the relative frame is the most common frame in standard varieties of English.

Besides the overall trend, the use of reference frames can vary with the circumstances. Related patterns have been studied in cognitive science.
For example, \citet{vukovic2015individual} find a three-way interaction between linguistic cues, spatial configurations in an image, and a person's own preferences on reference frames.

We investigated whether reference to a person in the image might influence how annotators comprehend the scene. 198 out of the 616 instances involve ``person'' in the caption. And out of the 198 human-involved instances, 32 (16\%) use an intrinsic frame and 154 (78\%) use a relative frame (12, i.e.~6\%, can be interpreted with both frames),
%while the proportions were 11.0\% and 84.1\% overall.
while the proportions were 9\% and 87\% for instances not involving ``person''.
%Even if we assume that ambiguous cases are all to be interpreted as the intrinsic frame, this is a statistically significant difference (using Fisher's exact test, two-tailed $p{=}0.052$).
This is a statistically significant difference
(using two-tailed Fisher's exact test,
$p{=}0.0054$ if ignoring both-frame cases,
and $p{=}0.0045$ if grouping both-frame and intrinsic cases).
In other words, this suggests that the involvement of a human can more likely prompt the use of the intrinsic frame.

%We argue that the task of identifying frame of reference itself is very hard for current models. However, learning to recognise frame of reference helps the task of visual spatial reasoning. We conduct a case study on the left/right-related relations by training a model to predict the human labelled reference frame given an image and a true statement. Secondly, we use the model trained with reference frame labels to initialise the VSR task model and further finetune the model on VSR left/right-related data points. Interestingly, we found frame of reference recognition helps the downstream task of VSR. We discuss more in \Cref{sec:exp_rf}.

% [TODO]: explain the link between allo-/ego-centric to reference frames.
% GE: It would take too long to explain. Levinson discusses various different terminologies

% Two paragraphs:
% 1: interesting variations (skewed towards..)
% 2: one source of variation is whether there is a human in the image

%%% quote from Vukovic and Williams 
%On the one hand, one could adopt an egocentric, or internal, perspective and imagine performing the action oneself. Equally, one could assume the role of an external observer of the action – in other words, an allocentric perspective.

% Fangyu: In other words, egocentric - intrinsic; allocentric - relative
%%%
\section{Experiments}\label{sec:exp}

In this section, we test VLMs on VSR. We first introduce baselines and experimental configurations in \Cref{sec:baseline}, then experimental results and analysis in \Cref{sec:main_results}. Then we discuss the role of frame of reference using experiments in \Cref{sec:exp_rf} and finally conduct sample efficiency analysis in \Cref{sec:sample_efficiency}.

\subsection{Baselines and Experiment Configurations}\label{sec:baseline}

\paragraph{Baselines.} For finetuning-based experiments, we test three popular VLMs: VisualBERT \citep{li2019visualbert}\footnote{\href{https://huggingface.co/uclanlp/visualbert-nlvr2-coco-pre}{huggingface.co/uclanlp/visualbert-nlvr2-coco-pre}}, 
LXMERT \citep{tan2019lxmert}\footnote{\href{https://huggingface.co/unc-nlp/lxmert-base-uncased}{huggingface.co/unc-nlp/lxmert-base-uncased}},
ViLT \citep{kim2021vilt}\footnote{\href{https://huggingface.co/dandelin/vilt-b32-mlm}{huggingface.co/dandelin/vilt-b32-mlm}}. All three models are stacked Transformers \citep{vaswani2017attention} that take image-text pairs as input. The difference mainly lies in how or whether they encode the position information of objects. We report only finetuned results but not direct inferences from off-the-shelf checkpoints since some of their pretraining objectives are inconsistent with the binary classification task of VSR, thus requiring additional engineering.

We additionally test the alt text pretrained dual-encoder CLIP \citep{radford2021learning} as an off-the-shelf baseline model (no finetuning).\footnote{\href{https://huggingface.co/laion/CLIP-ViT-H-14-laion2B-s32B-b79K}{huggingface.co/laion/CLIP-ViT-H-14-laion2B-s32B-b79K}} We follow \citet{clip_vsr} to construct negation or antonym of each individual relation. E.g., ``facing'' $\rightarrow$ ``facing away from'' and ``ahead of'' $\rightarrow$ ``not ahead of''. For each sample, we compare the embedding similarity of the image-caption pair and that of the negated caption. If the original pair has a higher probability then the model prediction is \texttt{True}, otherwise \texttt{False}. We call this method CLIP (w/ prompting). We only report direct prompting results without finetuning since CLIP finetuning is expensive.

%More details are given in \Cref{sec:main_results}.

%The exact hyperparameters and base pre-trained models used are listed in Appendix (\Cref{tab:hparams}).
%We implement all baselines using PyTorch and the huggingface library.

\paragraph{Experimental configurations.} 
We save checkpoints every 100 iterations and use the best-performing checkpoint on dev set for testing. All models are run three times using three random seeds. All models are trained with AdamW optimiser \citep{loshchilov2018decoupled}. The hyperparameters we used for training the three VLMs
%and the links to the initialisation weights of the four baseline models 
are listed in \Cref{tab:hparams}. 

\begin{table}[!ht] %[!htbp]
\small
%\setlength{\tabcolsep}{1.2pt}
%\vspace*{-0.3cm}
%\renewcommand{\arraystretch}{0.8}
\centering
\setlength{\tabcolsep}{2.3pt}
\begin{tabular}{lccccccc}
\toprule
model & lr &  batch size & epoch & token length \\
\midrule
%\cmidrule(lr){1-1} \cmidrule(lr){2-2}
VisualBERT & 2e-6 & 32 & 100 & 32 \\
LXMERT & 1e-5 & 32 & 100 & 32 \\
ViLT & 1e-5 & 12 & 30 & max \\
\bottomrule
% \toprule
% model & \multicolumn{4}{l}{Huggingface URL} \\
% %\cmidrule(lr){1-1} \cmidrule(lr){2-2}
% \midrule
% {VisualBERT} & \multicolumn{4}{l}{{\tiny \url{uclanlp/visualbert-nlvr2-coco-pre}}} \\
% {LXMERT} & \multicolumn{4}{l}{{\tiny \url{unc-nlp/lxmert-base-uncased}}} \\
% {ViLT} & \multicolumn{4}{l}{{\tiny \url{dandelin/vilt-b32-mlm}}} \\
% {CLIP} & \multicolumn{4}{l}{{\tiny \url{laion/CLIP-ViT-H-14-laion2B-s32B-b79K}}} \\
% \bottomrule
\end{tabular}
\caption{A listing of hyperparameters used for all VLMs (``lr'': learning rate).
%and model names of the four VLMs' pretrained weights used in this work.
}
\label{tab:hparams}
\end{table}

\subsection{Experimental Results}\label{sec:main_results}

\begin{figure*}[th!]
    \centering
\begin{subfigure}[b]{\linewidth}
    \centering
    \includegraphics[width=\linewidth]{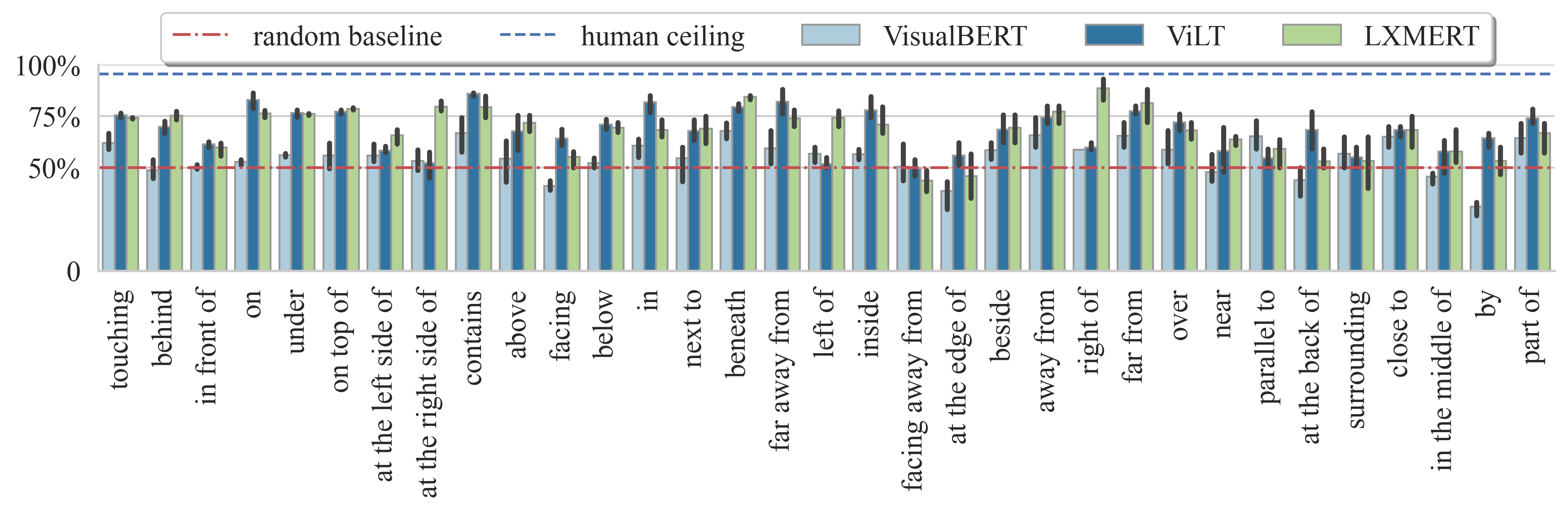}
    %\vspace{-1cm}
    \caption{random split}
\end{subfigure}
\begin{subfigure}[b]{\linewidth}
    \centering
    \includegraphics[width=\linewidth]{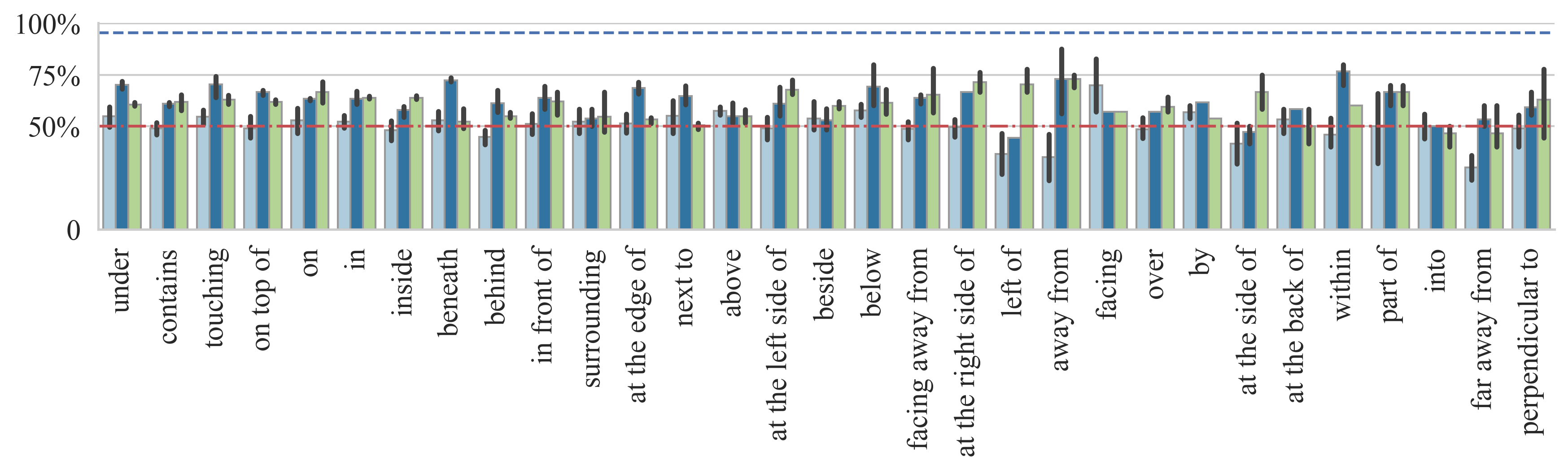}
    %\vspace{-1cm}
    \caption{zero-shot split}
\end{subfigure}
\caption{Performance (accuracy) by relation on the random (upper) and zero-shot (lower) split test sets. Relation order sorted by frequency (high to low from left to right). Only relations with more than 15 and 5 occurrences on the random and zero-shot tests respectively are shown. }
\label{fig:performance_by_rel}
%\vspace{-1em}
\end{figure*}

\begin{table}[ht]
\centering
\small
\begin{tabular}{lcccccc}
\toprule
model$\downarrow$ & random split & zero-shot split \\
\midrule
human ceiling & \multicolumn{2}{c}{95.4}   \\
\midrule
CLIP {\scriptsize(w/ prompting)} & 56.0 & 54.5 \\
\midrule
VisualBERT &  55.2$_{\pm1.4}$ & 51.0$_{\pm1.9}$  \\ 
%ViLT & 71.0$_{\pm0.7}$ & 62.4$_{\pm1.5}$ \\ 
ViLT & \textbf{69.3}$_{\pm0.9}$ &  \textbf{63.0}$_{\pm0.9}$  \\ 
LXMERT & \textbf{70.1}$_{\pm0.9}$ & 61.2$_{\pm0.4}$  \\ 
\bottomrule
\end{tabular}
\caption{Model performance on VSR test set. CLIP is applied without finetuning but with carefully engineered prompts while the other three smaller models are finetuned on the training set.}
%(both random and zero-shot splits are listed). }
\label{table:results}
\end{table}

In this section, we provide both quantitative and qualitative results of the four baselines. Through analysing the failure cases of the models, we also highlight the key abilities needed to solve this dataset.

As shown in \Cref{table:results}, the best-performing models on the random split are LXMERT and ViLT, reaching around 70\% accuracy while VisualBERT is just slightly better than the chance level. On the zero-shot split, all models' performance decline substantially and the best model ViLT only obtains 63.0\% accuracy.
The off-of-the-shelf CLIP model obtains around 55\% on both sets, indicating its weaknesses in spatial reasoning echoing \citet{subramanian2022reclip}'s findings.
Overall, these results lag behind the human ceiling by more than 25\% and highlight that there is very substantial room for improving current VLMs. 

\paragraph{Explicit positional information matters.} Both LXMERT and ViLT outperform VisualBERT by large margins ($>$10\%) on both splits.
This is expected since LXMERT and ViLT encode explicit positional information while VisualBERT does not. LXMERT has position features as part of the input which encodes the relative coordinates of objects within the image. ViLT slices an image into patches (instead of object regions) and uses positional encodings to signal the patches' relative positions.
VisualBERT, however, has no explicit position encoding. \citet{bugliarello-etal-2021-multimodal,positional2022} also highlight the importance of positional encodings of VLMs, which agrees with our observations.

\paragraph{Random split vs. zero-shot split.} 
It is worth noting that the performance gap between the random and zero-shot splits is large.
As we will show in~\cref{sec:sample_efficiency}, the underlying cause is not likely to be the number of training examples, but rather that concept zero-shot learning is fundamentally a challenging task.
The gap suggests that disentangling representations of concepts and relations is challenging for current models.

\paragraph{Sensitiveness to random seeds.}
%On both random and zero-shot splits' test sets, LXMERT is the most unstable model, with a standard deviation of 1.4 and 1.7 percentage points, respectively.
Model performance varies by about one to two percentage points.
% In general, the models have larger standard deviations on the zero-shot split, probably because the zero-shot dev/test sets are smaller.
%Due to the fluctuations we observe, we recommend
These fluctuations illustrate the importance of
always reporting the average performance of multiple runs to make sure the conclusion is reliable.

\begin{figure}[th!]
  \centering
        \centering
    \begin{subfigure}[b]{0.8\linewidth}
    \centering
    \includegraphics[width=\linewidth]{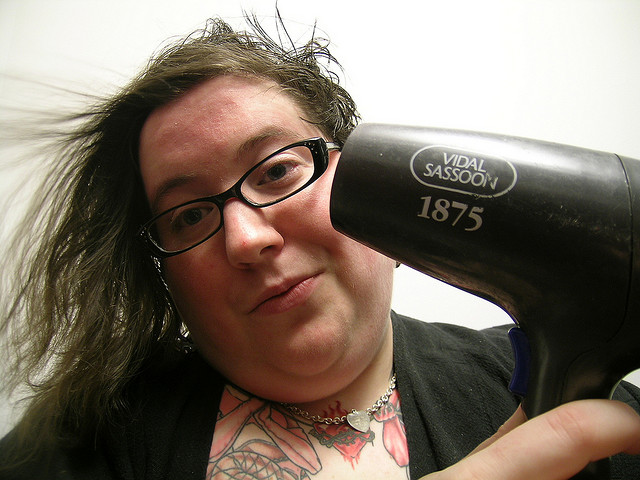}
    \caption{Caption: \textit{The hair drier is facing away from the person}. Label: \texttt{False}.}
     \end{subfigure}%
     \hfill
     \begin{subfigure}[b]{0.8\linewidth}
         \centering
    \includegraphics[width=\linewidth]{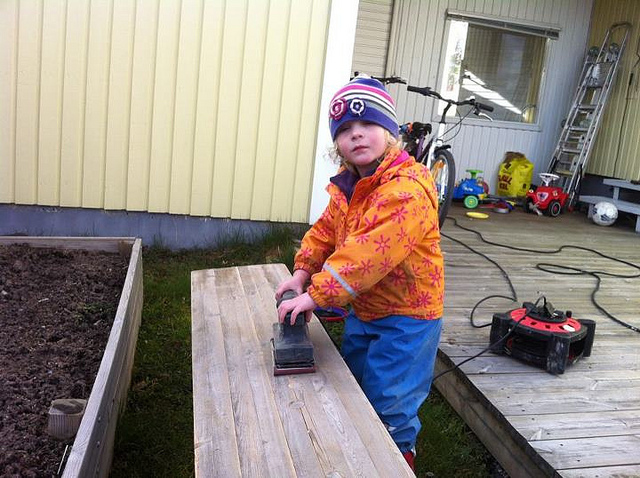}
         \caption{Caption: \textit{The bench is in front of the person}. Label: \texttt{True}.}
     \end{subfigure}
    
    \caption{LXMERT failed on both examples.}
    \label{fig:hair_drier_facing}
    %\vspace{-1em}
\end{figure}

\paragraph{Performance by relation.} We give performance by relation for all three finetuned models on both random and zero-shot splits in \Cref{fig:performance_by_rel}. The order from left to right is sorted by the frequency of relations in the dataset (within each split). Interestingly, there does not seem to be any correlation between performance and frequency of the relation, hinting that specific relations are hard not due to an insufficient number of training examples but because they are fundamentally challenging for current VLMs. Any relation that requires recognising orientations of objects seems to be hard, e.g., ``facing'', ``facing away from'', ``parallel to'' and ``at the back of''. As an example, LXMERT failed on the two examples in \Cref{fig:hair_drier_facing} which require understanding the front of a hair drier and a person respectively.
In this regard, left-right relations such as ``at the left/right side of'' and ``left/right of'' are difficult because the intrinsic reference frame requires understanding the orientation of objects. As an example, in \Cref{fig:potted_plant_bench}, all three models predicted \texttt{False}, but in the intrinsic frame (i.e., from the bench's point of view), the potted plant is indeed at the right.
%While generally speaking orientation has been hard, the tested VLMs do well on some seemingly hard cases. E.g., all three models correctly predicted \Cref{fig:fire_hydrant_has_face} -- a case that requires compositional zero-shot generalisation capability (to generalise the concept of ``face'' to a fire hydrant by identifying eyes).\footnote{We cannot entirely rule out the possibility that models relied on other spurious cues though.}

% \begin{figure}[]
%   \centering
%     \includegraphics[width=0.4\linewidth]{images/000000147270.jpg}
%     \caption{Caption: \textit{The fire hydrant is facing away from the person}. Label: \texttt{True}. All three models predicted correctly.}
%     \label{fig:fire_hydrant_has_face}
% \end{figure}

\begin{figure}[t]
    \centering
    \begin{subfigure}[b]{\linewidth}
    \centering
    \includegraphics[width=0.96\linewidth]{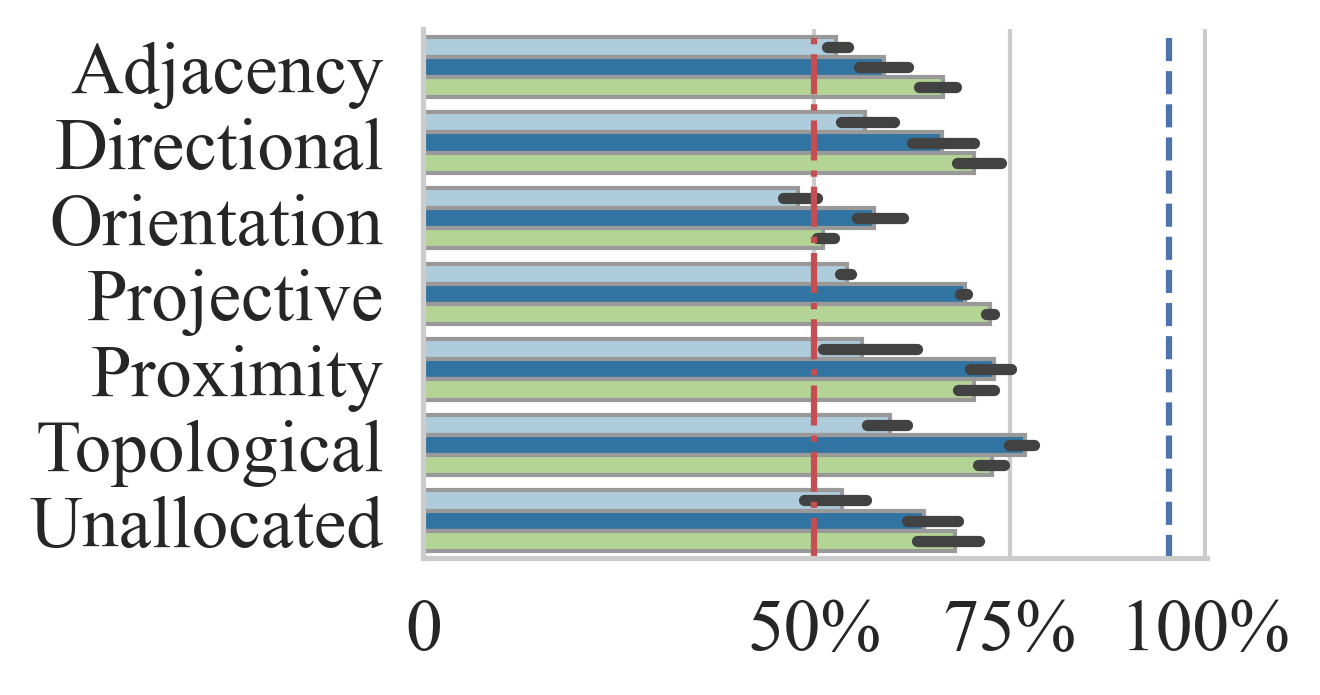}
    \caption{random split}
     \end{subfigure}
     \begin{subfigure}[b]{\linewidth}
         \centering
    \includegraphics[width=0.96\linewidth]{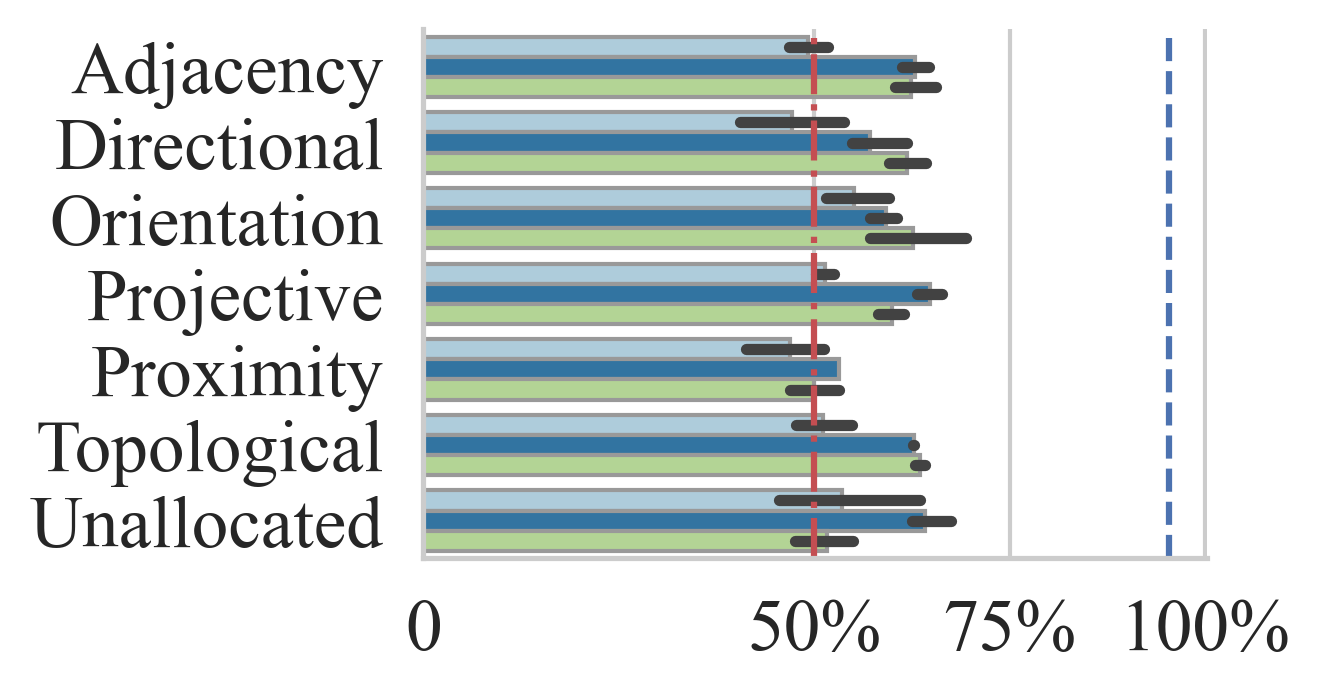}
    \caption{zero-shot split}
     \end{subfigure}
\caption{Performance by categories of relations, on the random and zero-shot split test sets. For legend information, see \Cref{fig:performance_by_rel}.}
\label{fig:performance_by_meta_cat}
    %\vspace{-1em}
\end{figure}

To get a more high-level understanding of the relations' performance, we group model performance by the categories of \citet{marchi2021cross}: ``Adjacency'', ``Directional'', ``Orientation'', ``Projective'', ``Proximity'', ``Topological'' and ``Unallocated'' (also shown in \Cref{tab:spatial_relations}). The results are shown in \Cref{fig:performance_by_meta_cat}. ``Orientation'' is the worst performing group on the random split, and on average all models' performances are close to the chance level. When comparing random and zero-shot splits, performance has declined to some extent for almost all categories and models.
%(with ViLT suffered the least in ``Topological'' ). 
The decrease in ``Proximity'' is particularly drastic across all models -- it declined from close to 75\% accuracy in random split to chance level in zero-shot split. ``Proximity'' contains relations such as ``close to'', ``near'' and ``far from''. We believe it is due to the fact that the notion of proximity is relative and very much dependent on the nature of the concept and its frequent physical context. E.g., for a ``person'' to be ``near'' an indoor concept such as ``oven'' is very different from a person being ``near'' a frequent outdoor object such as ``train'' or ``truck''. Since the zero-shot split prevents models from seeing test concepts during training, the models have a poor grasp of what counts as ``close to'' or ``far from'' for these concepts, thus generalising poorly.

%TODO: use a concrete example

\begin{figure}[!tbp]
  \centering
    \includegraphics[width=0.8\linewidth]{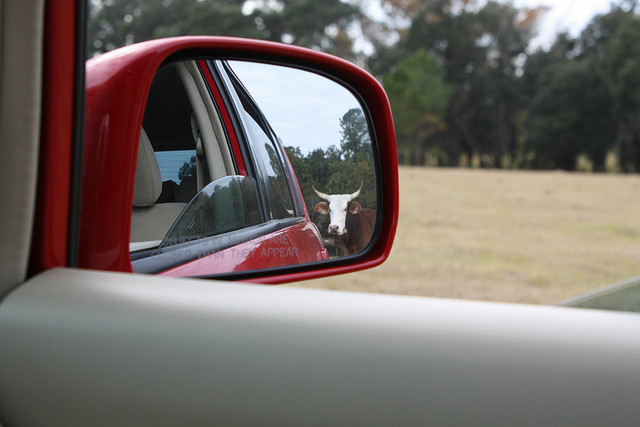}
    \caption{Caption: \textit{The cow is at the back of the car}. Label: \texttt{True}. LXMERT and VisualBERT predicted \texttt{False}.}
    \label{fig:cow_back_car}
    %\vspace{-1em}
\end{figure}

\paragraph{Other Errors.} While certain relations are intrinsically hard, we have observed other types of errors that are not bounded to specific relations. Here we give a few examples. Some instances require complex reasoning. In \Cref{fig:cow_back_car}, the model needs to recognise that both the cow and the back of the car are in the car's side mirror and also infer the relative position of the back of the car and the cow. It is perhaps no surprise that two of the three models predicted wrongly. Some other examples require common sense. E.g., in \Cref{fig:person_cow}, we can infer the person and the cow's moving direction and can then judge if the cow is ahead of the person. 
LXMERT failed on this example. 
In \Cref{fig:annotation_instance} (right), the model needs to infer that the main body of the cat is hidden behind the laptop. Interestingly, all three models predicted this example correctly.

% \begin{figure}[!tbp]
%   \centering
%     \includegraphics[width=0.7\linewidth]{images/000000119360.jpg}
%     \caption{Caption: \textit{The cat is behind the laptop}. Label: \texttt{True}.}
%     \label{fig:cat_behind_laptop}
% \end{figure}

\begin{figure*}[!thp]
    \centering
    \includegraphics[width=\linewidth]{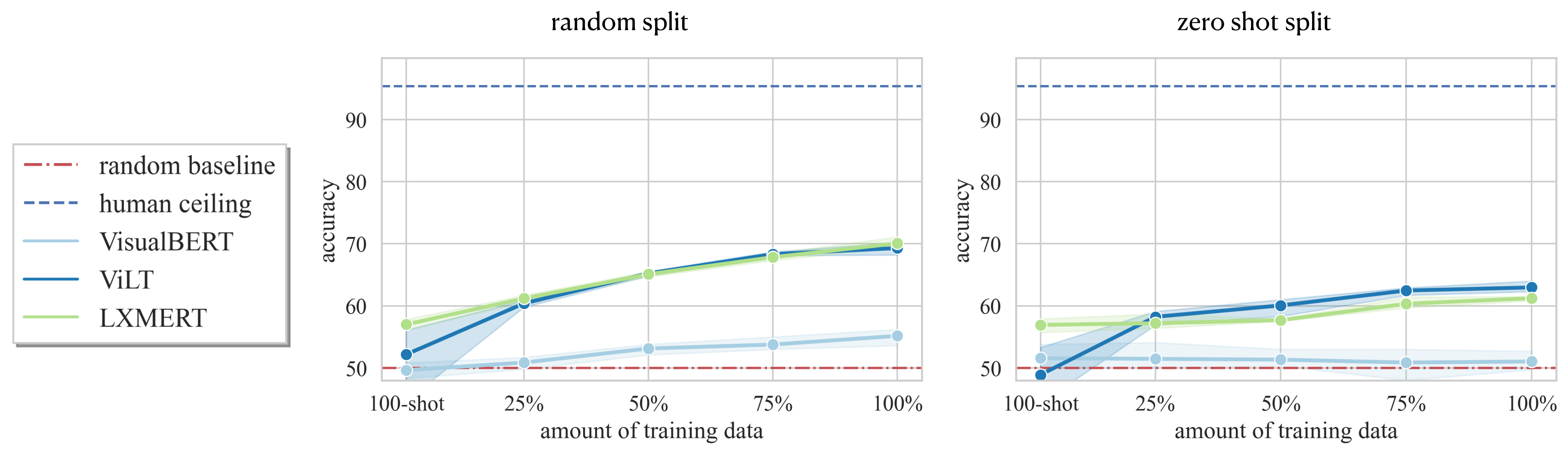}
\caption{Sample efficiency analysis: model performance under different amounts of training data (100-shot, 25\%, 50\%, 75\% and 100\% of training set). Results on both the random and zero-shot split test sets are shown. As training data increases, the performance plateaus on both sets but the flattening trend is more obvious on the zero-shot split.}
    \label{fig:sample_efficiency}
    %\vspace{-1em}
\end{figure*}

\subsection{Case Study on Reference Frames}\label{sec:exp_rf}

As discussed in \Cref{sec:relative_vs_intrinsic}, different frames of reference can be used in natural language and it would be helpful to understand whether our models recognise them.
We argue that the task of identifying frame of reference itself is very hard for current models. However, learning to recognise frames of reference helps the task of visual spatial reasoning.

Firstly, we conduct a case study on left/right-related relations. We additionally label the reference frames of all true statements containing any of the left/right-related relations. We exclude all data points that can be interpreted in both intrinsic and relative frames to slightly reduce the complexity of the task.
Then we finetune a ViLT checkpoint to predict the reference frame based on the true statement and the image. The model's performance on test set is shown in the upper half of \Cref{table:rf}. We can see that reference frame prediction is an extremely hard task for the model. This is presumably because it requires taking into account a 3D viewpoint and simulating transformations between different viewpoints.

Secondly, we use this model trained with reference frame labels to initialise the VSR task model and further finetune it on the VSR task (only the left/right relations). The test results are shown in the lower part of \Cref{table:rf}.\footnote{Note that the reference frame train/dev/test sets are derived from the VSR task split -- so no data leakage is possible from train to dev and test sets even after the intermediate pretraining.} We see a clear positive transfer from reference frame prediction task to the VSR task.
%especially on the left/right-relation setup. 
This suggests that learning to recognise reference frames can indeed help downstream visual spatial reasoning. This makes sense since simulating the transformation of intrinsic/relative frames could be an intermediate reasoning step in detecting whether a statement is true/false.

\begin{table}[]
\centering
\small
\begin{tabular}{lccccc}
\toprule
\multicolumn{4}{c}{\emph{Reference frame prediction task}} \\
\midrule
model$\downarrow$ & Precision & Recall & F1  \\
\midrule
ViLT & 59.2$_{\pm3.7}$ & 59.7$_{\pm5.8}$ & 56.9$_{\pm4.4}$ \\
\toprule
\multicolumn{4}{c}{\emph{VSR task (left/right subset)}} \\
\midrule
model$\downarrow$ &  & \multicolumn{2}{c}{Accuracy} \\
\midrule
\multicolumn{2}{l}{ViLT} & \multicolumn{2}{c}{54.2$_{\pm0.6}$} \\
\multicolumn{2}{l}{ViLT + rf\_trained} &  \multicolumn{2}{c}{\textbf{59.2}$_{\pm1.8}$} \\
\bottomrule
\end{tabular}
\caption{ViLT model performance on the reference frame prediction task (upper half; we report macro-averaged Precision/Recall/F1 since the binary classification task is imbalanced); and VSR task using original pretrained checkpoint or the reference frame prediction task trained checkpoint (accuracy reported).}
\label{table:rf}
\end{table}

\subsection{Sample Efficiency}\label{sec:sample_efficiency}
In order to understand the correlation between model performance and the number of training examples, we conduct sample efficiency analysis on VSR. The results are plotted in \Cref{fig:sample_efficiency}. For the minimum resource scenario, we randomly sample 100 shots from the training sets of each split. Then we gradually increase the number of training examples to be 25\%, 50\% and 75\% of the whole training sets. 
Both LXMERT and ViLT have a reasonably good few-shot capability and can be quite performant with 25\% of training data. LXMERT, in particular, reaches above 55\% accuracy with 100 shots on both splits.
The zero-shot split is substantially harder and most models appear to have already plateaued at around 75\% of the training set. For the random split, all models are increasing performance with more data points, though improvement slows down substantially for LXMERT and ViLT after 75\% of training data. The fact that LXMERT has the best overall few-shot capability may be suggesting that LXMERT's pretrained object detector has a strong inductive bias for the VSR dataset as it does not need to learn to recognise concept boundaries and classes from scratch. However, this advantage from LXMERT seems to fade away as the number of training examples increases. 

% \begin{figure}
%     \centering
%     \begin{subfigure}[b]{0.49\linewidth}
%     \centering
%     \includegraphics[width=\linewidth]{images/vsr_sample_efficiency_dec_2.pdf}
%     \caption{random split}
%      \end{subfigure}
%      \begin{subfigure}[b]{0.49\linewidth}
%          \centering
%     \includegraphics[width=\linewidth]{images/sample_efficiency_zeroshot_split_v2.png}
%     \caption{zero-shot split}
%      \end{subfigure}

%\subsection{Case Studies} % do case studies along the way

\section{Conclusion and Future Directions}\label{sec:conclusion}
We have presented Visual Spatial Reasoning (VSR), a controlled probing dataset for testing vision-language models (VLMs)' capabilities of recognising and reasoning about spatial relations in natural image-text pairs. 
We made a series of linguistic observations on the variability of spatial language when collecting VSR. We highlighted the diverse use of reference frames among annotators, and also the ambiguous nature of certain spatial relations. 
%We ensure the quality of the dataset by 
We tested four popular VLMs on VSR, and found they perform more than 25\% below the human ceiling. 
On a more challenging concept zero-shot split, the tested VLMs struggled to reach 60\% accuracy and their performance plateaued even with increased training examples. 
Among the finetuning-based VLMs, ViLT and LXMERT outperformed VisualBERT, and we pointed out that the explicit positional information in the former two models is crucial in the task. CLIP with prompt engineering achieved slightly better than random performance, suggesting poor capability in spatial reasoning.
We also performed a by-relation analysis and found that the models' performances on certain relations have little correlation with the number of training examples, and certain relations are inherently more challenging. 
We identified orientation as the most difficult category of relations for VLMs. Proximity is another challenging category, especially in the zero-shot setup as this relation is highly concept-dependent. 
We hope the task serves as a useful tool for testing and probing future VLMs.

In future work, we plan to more extensively investigate whether large-scale pretrained dual-encoders such as CLIP \citep{radford2021learning}, ALIGN \citep{jia2021scaling} and LiT \citep{zhai2021lit} can properly recognise spatial relations, especially in the finetuning setup. 
A comparison of dual- and cross-encoders' performance on each spatial relation might guide future model design.
Recently, \citet{flamingo2022,chen2023pali,huang2023language} proposed ultra-large-scale VLMs. It would be interesting to see if VLMs have better spatial reasoning capability when scaled up.
Another direction is extending VSR to cover more languages and cultures \citep{liu-etal-2021-visually,bugliarello2022iglue} and test multilingual VLMs. 
Along the same line, since we have also collected the metadata of annotators, the VSR corpus can be used as a resource for investigating research questions such as: How is ``space'' described among different dialects of English? How is ``space'' perceived among different populations? We hope that the annotation process of VSR can also serve as a basis for future cross-lingual and cross-cultural sociolinguistic research.

\section*{Acknowledgements}
%\section{Acknowledgements}

We thank the TACL reviewers and the action editor for their thoughtful comments. We thank Qian Wang and Rongtian Ye for helping trial the annotation scheme; Zihao Fu for helping set up the annotation server.
The project is funded by Cambridge Language Sciences Incubator Fund. FL is supported by Grace \& Thomas C.H. Chan Cambridge Scholarship.

\bibliography{anthology,custom}
\bibliographystyle{acl_natbib}

\newpage
\appendix
\section{Appendix}
\label{sec:appendix}

\begin{figure}[!htbp]
  \centering
    \includegraphics[width=0.6\linewidth]{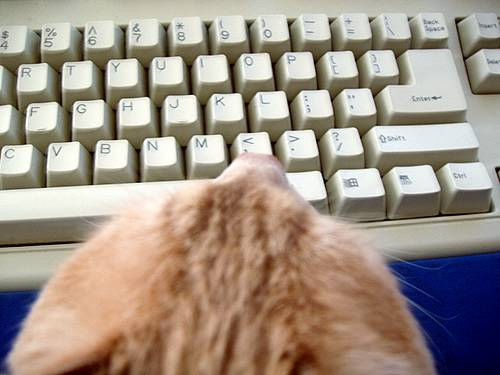}
    \caption{Caption: \textit{The keyboard is below the cat}.}
    \label{fig:ambiguity_1}
\end{figure}

\paragraph{Genuinely ambiguous annotation instances.} As mentioned in the main text, we analysed 100 examples with high disagreement among annotators and found that the majority of the case are truly ambiguous and cannot be assigned a definitive True/False label. Here we show two concrete examples. 
In \Cref{fig:ambiguity_1}, we can infer that the keyboard is horizontally lower than the cat. However, the keyboard is not right under the cat's body (but a bit in front of the cat). The ambiguity results from the semantics of the word ``below'' which can mean both (1) horizontally lower (but not necessarily right under) and also (2) right under. 
In \Cref{fig:ambiguity_2}, if seat surface is viewed as the main body of the chair, we can infer that the sandwich is indeed above the chair. However, it is not above all parts of the chair (it is no higher than the backrest of the chair). This case depends on a person's subjective understanding of what counts as ``above the chair'' and is thus also intrinsically ambiguous.

\begin{figure}[!htbp]
  \centering
    \includegraphics[width=0.6\linewidth]{images/000000044682.jpg}
    \caption{Caption: \textit{The sandwich is above the chair}.}
    \label{fig:ambiguity_2}
\end{figure}

While we have excluded such cases in the standard VSR train/val/test split to ensure the dataset has human consensus, we also release the raw dataset without excluding these ambiguous cases for researchers who are interested in studying them further.

% \paragraph{Model performance on dev sets.} For transparency, we also list the models' dev set performances in \Cref{table:full_results}. The gap between dev and tests become much greater on zero-shot split likely due to the size of both dev and test sets have become smaller, making the evaluation less stable.

% \begin{table}[ht]
% \centering
% \setlength{\tabcolsep}{2.8pt}
% \small
% \begin{tabular}{lcccccc}
% \toprule
% & \multicolumn{2}{c}{random split} &  & \multicolumn{2}{c}{zero-shot split} \\
% \cmidrule(l){2-3} 	\cmidrule(l){4-6}
% model$\downarrow$ & dev & test & & dev & test  \\
% \midrule
% human & \multicolumn{5}{c}{95.4}   \\
% \midrule
% VisualBERT & 59.2$_{\pm0.9}$ & 57.4$_{\pm0.9}$ & & 57.4$_{\pm2.2}$  & 54.0$_{\pm1.3}$  \\ 
% LXMERT & \textbf{73.8}$_{\pm1.2}$  & \textbf{72.5}$_{\pm1.4}$ & & \textbf{69.2}$_{\pm1.0}$  & \textbf{63.2}$_{\pm1.7}$  \\ 
% ViLT & 71.9$_{\pm1.3}$  & 71.0$_{\pm0.7}$  & & 66.7$_{\pm1.7}$  & 62.4$_{\pm1.5}$ \\ 
% \bottomrule
% \end{tabular}
% \caption{Model performance on VSR. Results of both random and zero-shot splits, both validation and tests are listed. (Dev performance using the best checkpoint on dev set.)}
% \label{table:full_results}
% \end{table}

\paragraph{Relations included and excluded.} Our used set of relations is adapted from \citet{marchi2021cross}. In \Cref{tab:spatial_relations_original}, we copy their original table for reference. We also list the exact modifications we made on top of the original table in \Cref{tab:spatial_relations_adaptations}. The major reasons for excluding relations are: (1) rarely used in describing spatial information of images (e.g., ``north of''), (2) repeated with other relations (e.g., ``front of'' and ``in front of''). In order to convert relations into templates, we also inserted some words to make them grammatical. We included several additional relations since we found them very frequent in MS COCO (e.g., ``at the edge of'' and ``touching'').

\begin{table*}[]
\small
    \centering
    \begin{tabular}{l|l}
    \toprule
        \rowcolor{DarkGray}
    Category & Spatial Relations \\
    \midrule
    Adjacency     & Adjacent, alongside, side, right side, left side, attached, back side, front of, ahead, against \\
        \rowcolor{Gray}
    Cardinal direction & North of, south side of, east-west, easterly, north-south, south of \\
 Directional & \makecell[l]{Off, past, toward, down, deep down, up, away from, along, around, from, into, to, across, \\across from, through, down from }\\
    \rowcolor{Gray}
    Orientation & Facing, orientation, parallel, perpendicular \\
    Projective & \makecell[l]{On top of, beneath, in back of, beside, behind, left of, right of, under, in front of, below,\\ above, over}\\
        \rowcolor{Gray}
    Proximity & By, close to, near, far, far from, far away from \\
    Topological & \makecell[l]{Congruent, connected, detached, has part, part of, contains, within, at, on, in, with, surround,\\ among, consists of, out of, between, inside, outside}\\
        \rowcolor{Gray}
    Unallocated & Beyond, next to, opposite, after, among, enclosed by \\
\bottomrule
    \end{tabular}
    \caption{The original table from \citet{marchi2021cross}.}
    \label{tab:spatial_relations_original}
\end{table*}

\begin{table*}[]
\small
    \centering
    \begin{tabular}{l|l}
    \toprule
        \rowcolor{DarkGray}
    Category & Spatial Relations \\
    \midrule
    Adjacency     & \makecell[l]{Adjacent \textcolor{blue}{to}, alongside, \textcolor{blue}{at the} side \textcolor{blue}{of}, \textcolor{blue}{at the} right side \textcolor{blue}{of}, \textcolor{blue}{at the} left side \textcolor{blue}{of}, attached \textcolor{blue}{to},\\ \textcolor{blue}{at the} back \textcolor{red}{side} \textcolor{blue}{of}, \textcolor{red}{front of}, ahead \textcolor{blue}{of}, against, \textcolor{blue}{at the edge of}} \\
        \rowcolor{Gray}
    Cardinal direction & \textcolor{red}{North of}, \textcolor{red}{south side of}, \textcolor{red}{east-west}, \textcolor{red}{easterly}, \textcolor{red}{north-south}, \textcolor{red}{south of} \\
 Directional & \makecell[l]{Off, past, toward, down, deep down, up, away from, along, around, from, into, to, across, \\across from, through, down from }\\
    \rowcolor{Gray}
    Orientation & Facing, \textcolor{blue}{facing away from}, \textcolor{red}{orientation}, parallel \textcolor{blue}{to}, perpendicular \textcolor{blue}{to}\\
    Projective & \makecell[l]{On top of, beneath, \textcolor{red}{in back of}, beside, behind, left of, right of, under, in front of, below,\\ above, over, \textcolor{blue}{in the middle of}}\\
        \rowcolor{Gray}
    Proximity & By, close to, near, \textcolor{red}{far}, far from, far away from \\
    Topological & \makecell[l]{\textcolor{red}{Congruent}, connected \textcolor{blue}{to}, detached \textcolor{blue}{from}, has \textcolor{blue}{as a} part, part of, contains, within, at, on, in, with,\\ surround\textcolor{blue}{ing}, among, consists of, out of, between, inside, outside, \textcolor{blue}{touching}}\\
        \rowcolor{Gray}
    Unallocated & Beyond, next to, opposite \textcolor{blue}{to}, after, among, enclosed by \\
\bottomrule
    \end{tabular}
    \caption{Our used table. \textcolor{blue}{What we added}. \textcolor{red}{What's deleted}.}
    \label{tab:spatial_relations_adaptations}
\end{table*}

\begin{figure*}[!tbp]
  \centering
    \includegraphics[width=0.98\linewidth]{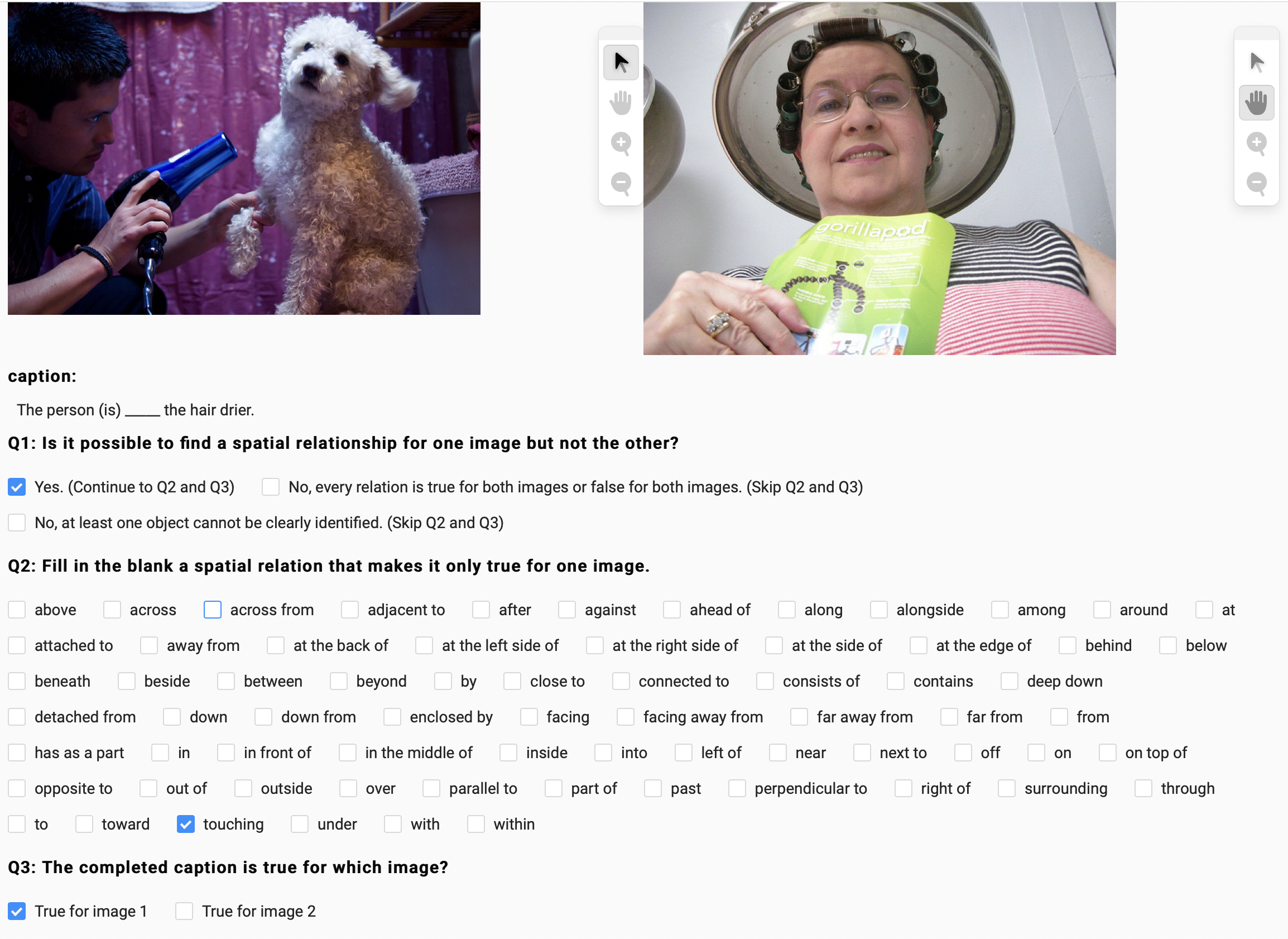}
    \caption{A screenshot of the annotation interface.}
    \label{fig:annotation_interface}
\end{figure*}

\begin{figure*}[!tbp]
  \centering
    \includegraphics[width=0.6\linewidth]{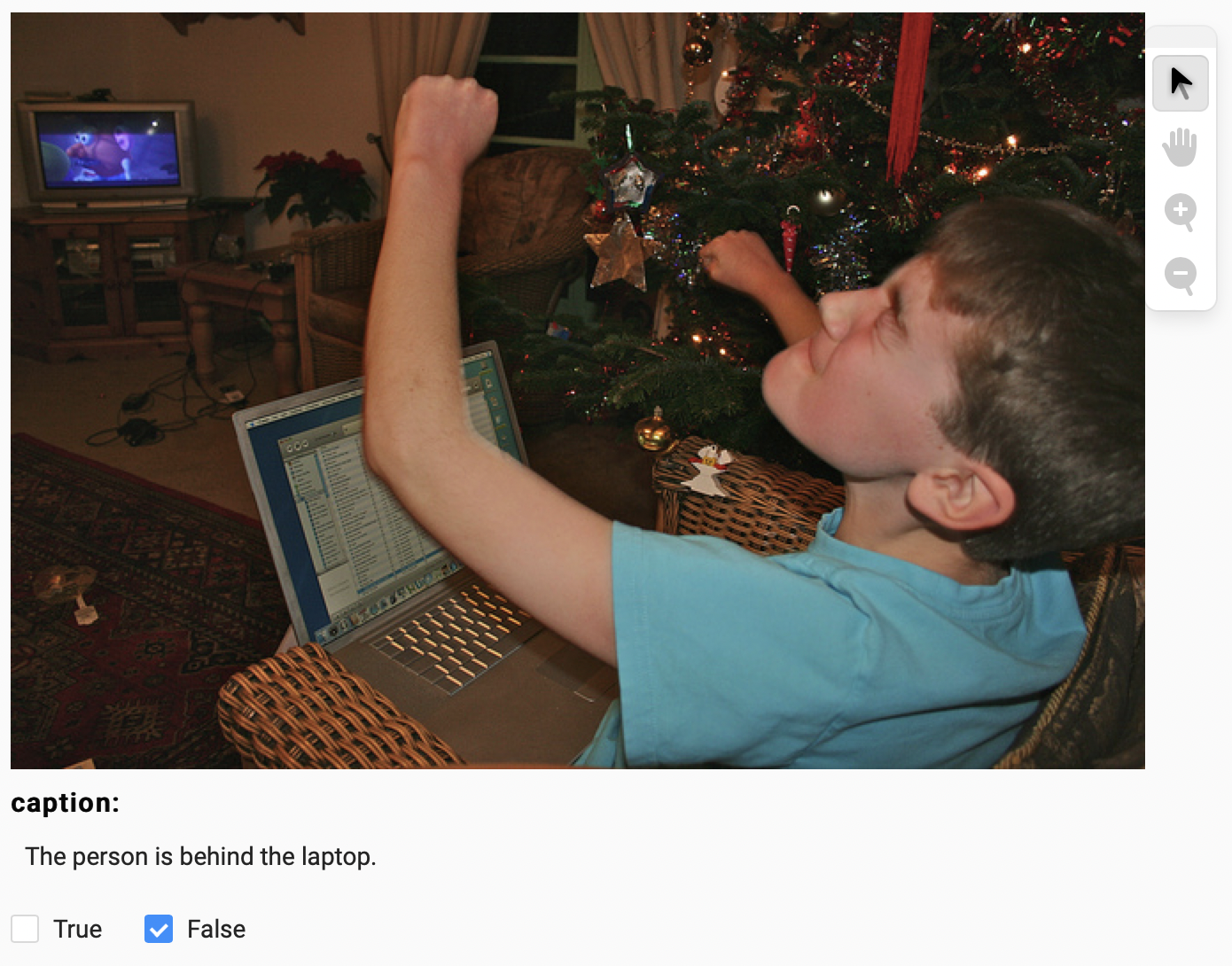}
    \caption{A screenshot of the validation interface.}
    \label{fig:validation_interface}
\end{figure*}

\paragraph{Screenshots of the annotation interface.} We use the open-sourced label studio (\url{labelstud.io}) for managing our annotation tasks. Two screenshots are shown in \Cref{fig:annotation_interface} (caption generation) and \Cref{fig:validation_interface} (validation).

\end{document}